\documentclass{article}
\usepackage{arxiv}

\usepackage{newtxtext,newtxmath}

\usepackage{graphicx}
\usepackage{algorithm}
\usepackage{algpseudocode}
\usepackage{makecell}
\usepackage{setspace}
\usepackage{microtype}      

\usepackage{doi}
\usepackage{float}
\usepackage{afterpage}
\usepackage{caption}

\usepackage{siunitx}
\usepackage{wasysym}
\usepackage{titling}

\usepackage{xcolor}

\definecolor{headingcolor}{HTML}{792F3A}

\usepackage{titlesec}

\titleformat{\section}
  {\Large\bfseries\raggedright\color{headingcolor}}
  {\thesection}{1em}{}

\titleformat{\subsection}
  {\large\bfseries\raggedright\color{headingcolor}}
  {\thesubsection}{1em}{}

\titleformat{\subsubsection}
  {\normalsize\itshape\color{headingcolor}}
  {\thesubsubsection}{1em}{}

\titlespacing*{\section}{0pt}{12pt}{-2pt}

\titlespacing*{\subsection}{0pt}{10pt}{-2pt}

\titlespacing*{\subsubsection}{0pt}{8pt}{-2pt}

\date{}

\renewenvironment{abstract}
	{\quotation}
	{\endquotation}

\date{}

\makeatletter
\renewcommand{\fnum@figure}{\textbf{Figure \thefigure}}
\renewcommand{\fnum@table}{\textbf{Table \thetable}}
\makeatother

\usepackage[numbers,sort&compress]{natbib}

\usepackage{url}
\hypersetup{
    colorlinks=true,
    urlcolor=blue,
    linkcolor=blue,
    citecolor=blue,
    filecolor=blue
}

\def\scititle{
	The Missing Touch: Sensing and Displaying Localized Contact Geometry Brings Teleoperation Closer to Human Dexterity
}
\title{\bfseries \boldmath \scititle}

\author{
	Rohan Kota,
	Gregory Reardon$^{\dagger}$,
	J. Edward Colgate$^{\ast}$$^{\dagger}$\and
	\small Department of Mechanical Engineering, Center for Robotics and Biosystems,\and
    \small Northwestern University, Evanston, IL, USA. \\
	\small$^\ast$Corresponding author. Email: colgate@northwestern.edu \\
    \small$^\dagger$These authors contributed equally to this work.
}

\begin{document}

\captionsetup[figure]{font={scriptsize,sf}}
\captionsetup[table]{font={scriptsize,sf}}

\sisetup{inter-unit-product = \ensuremath{{\cdot}}} 

\maketitle

\begin{abstract} \bfseries \boldmath
A fundamental challenge in robotic teleoperation is enabling an operator to control a remote robot as effortlessly and intuitively as their own hands. Despite the growing use of teleoperation to collect demonstrations for training autonomous robot policies, teleoperated performance still falls significantly short of human dexterity, even for basic tasks. Here, we present evidence that a key factor contributing to this performance gap is the absence of localized tactile feedback. We equip a two-degree-of-freedom (DoF) bilateral force-feedback telemanipulator with a vision-based tactile sensor and a 32-pixel tactile display, allowing us to capture spatial contact patterns on the robot finger and render them on the operator’s fingertip. In teleoperated button-discrimination and peg-rolling tasks, we demonstrate that reproducing spatially distributed contact information brings teleoperated movements closer to natural human behavior through fewer unnecessary motions and task completion steps, significantly reducing task completion time and producing more consistent demonstrations across trials and operators. We further show that these improvements in human demonstrations can propagate to autonomous policies: policies trained on tactile-enhanced demonstrations exhibited similar reductions in both task completion time and deviation from natural human behavior. Together, these results establish spatially distributed tactile feedback as an important component of dexterous robotic teleoperation and demonstrate its potential to improve safety-critical teleoperation and enable more humanlike demonstrations for training autonomous robot policies.
\end{abstract}

\begin{figure}
	\centering
	\includegraphics[width=0.8\textwidth]{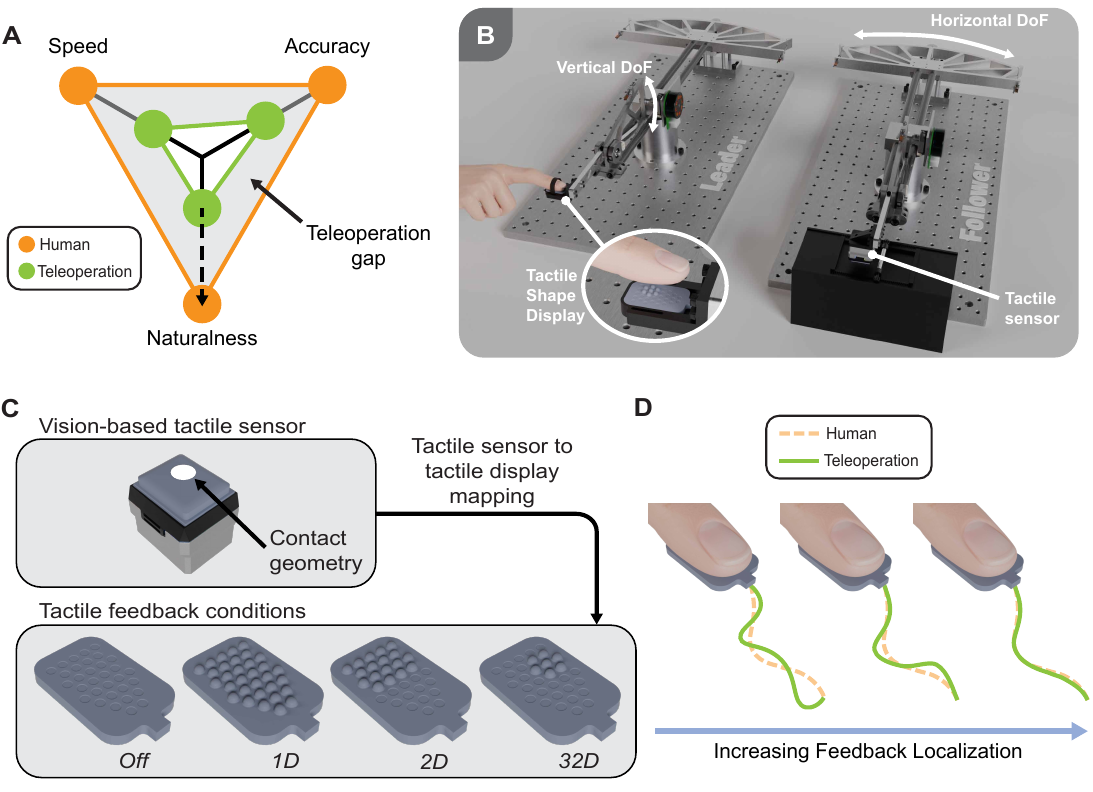}

    \singlespace
	\caption{\textbf{Spatially distributed cutaneous feedback improves the naturalness of teleoperated motion.}
		(\textbf{A}) Humans performing tasks through a robotic teleoperation system (green) move more slowly, less accurately, and less naturally than during direct manipulation (orange). Although prior works demonstrate that haptic feedback reduces the teleoperation gap in speed and accuracy, closing it fully requires restoring natural movement patterns (dashed arrow). (\textbf{B}) Participants used a 2-degree-of-freedom (DoF) leader-follower teleoperation device with bilateral kinesthetic force feedback, driven by brushless DC motors and capstan transmissions. The device allows operators to move in the vertical and horizontal directions. A vision-based tactile sensor serving as the robot ``finger" records spatially distributed contact during interaction with the environment, and a 32-pixel tactile shape display reproduces this distributed contact information on the operator's fingertip via controlled inflation of small elastic domes. (\textbf{C}) Contact patterns on the vision-based tactile sensor (top panel) are mapped to patterns on the tactile display (bottom panel). Bottom panel illustrates the four experimental feedback conditions: \textit{Off}: no cutaneous feedback; \textit{1D}: uniform inflation across the display upon contact; \textit{2D}: independent inflation of the upper and lower display halves based on contact location; \textit{32D}: localized inflation matching the distribution of contact on the robot finger. (\textbf{D}) Teleoperated movement patterns converge toward natural human movements as the localization of tactile feedback improves. Trajectories are illustrative and do not represent experimental measurements.
	}
	\label{fig:overview}
\end{figure}

\noindent

\section*{INTRODUCTION}
When a person reaches for a cup, they produce a movement of striking regularity: their hand follows a smooth, bell-shaped velocity profile~\cite{flash, morasso, vaisman}, settles onto the handle without overshoot, and grips with just enough force to prevent slip~\cite{johansson}. Human manipulation is fast, efficient, and robust to external disturbances, avoiding large unnecessary movements and exhibiting low variability and failure rates. A key contributor to this performance is the spatially distributed nature of touch~\cite{delhaye}, which provides continuous information about contact location, pressure distribution, and incipient slip at the fingertips. When this cutaneous information is removed by anesthetizing the fingertips, however, manipulation degrades sharply: people grip objects much harder than necessary, drop them more frequently, and rely increasingly on vision to compensate~\cite{johansson_glabrous, westling, augurelle, jenmalm, ung}.

Robotic teleoperation~\cite{hokayem, zhang, zhao, qin} has become a major pathway for transferring human manipulation skills to robots~\cite{argall, jaquier, zhao, fu, chi}. Yet despite decades of work on force feedback during robotic teleoperation~\cite{hannaford, syeda, lawrence, niemeyer}, without spatially resolved cutaneous cues at the fingertip, operators experience a sensory deficit similar to fingertip anesthesia: they feel interaction forces but not the local contact patterns that normally guide manipulation. As a result, they exhibit many of the same compensatory behaviors observed under cutaneous anesthesia~\cite{jenmalm, ung}, producing slower~\cite{pacchierotti, pacchierotti2016, meli, massimino, tavakoli} and less accurate~\cite{okamura} demonstrations with more unnecessary movements than direct manipulation (Fig.~\ref{fig:overview}A). Autonomous policies trained on such demonstrations inherit these inefficiencies, executing tasks more slowly and less accurately, along trajectories that depart from the smooth and efficient motions characteristic of direct human manipulation. Improving the naturalness of teleoperated demonstrations can therefore improve the quality of the data from which autonomous policies learn.

Rather than reproducing distributed contact patterns at the fingertip, many researchers have used the cutaneous channel to render low-dimensional force information through vibrotactile or electrotactile stimulation~\cite{massimino, tavakoli, okamura}, skin-stretch displays~\cite{schorr, saudrais}, or uniformly tilting platforms mounted on the fingertip~\cite{pacchierotti, pacchierotti2016, meli}. Although rendering force information through the cutaneous channel has been shown to improve task speed, accuracy, and force regulation~\cite{pacchierotti, meli, okamura, portoles}, these outcome measures provide limited insight into how haptic feedback influences gross movement patterns during teleoperation. Movement-based metrics, in contrast, can reveal differences in dexterity that may not be apparent from task-level performance alone and have been widely used to assess human skill in surgical tasks~\cite{thijssen, reiley, vedula}, including robotic surgery~\cite{reiley, okamura, sharon, wagner, moody}. However, the extent to which teleoperated movements deviate from natural human movement patterns---and the factors that might reduce these deviations---remains poorly understood.

Here, we investigate whether reproducing spatially distributed contact information can make teleoperated behavior more closely resemble natural human manipulation. Modern robotic hands increasingly integrate high-resolution tactile sensors, such as vision-based sensors~\cite{johnson, johnson_cole, shimonomura}, which have been shown to improve autonomous robot performance in contact-rich tasks~\cite{funk, george, calandra, yamaguchi, huang}. However, these rich tactile measurements are rarely conveyed to human operators during teleoperation, potentially limiting their ability to adapt their movements to contact conditions and generate higher-quality demonstrations for downstream policy learning~\cite{cuan, tan}. Although prior teleoperation systems have rendered robot-side contact information through the cutaneous channel~\cite{pacchierotti, pacchierotti2016, meli} and demonstrated improvements in task performance, these approaches effectively collapse a spatially distributed signal into a low-dimensional cue. Preserving the spatial structure of tactile information instead requires displays capable of producing localized, independently controllable deformations across the fingertip. Spatially distributed cutaneous displays, such as pin arrays~\cite{wagner_pinarray}, offer a natural approach for doing so, but achieving high spatial resolution typically requires a large number of actuators, resulting in devices that are bulky and difficult to integrate into teleoperation systems. Recent advances in actuator technology, such as embedded electro-osmotic pumps (EEOPs)~\cite{shultz, shen} and dielectric elastomers~\cite{koo, youn}, offer opportunities to overcome this trade-off, enabling compact tactile displays with spatial resolutions approaching the roughly 2-mm two-point discrimination threshold of the human fingerpad~\cite{weinstein}.

In this work, we integrate a compact, high-resolution tactile shape display (Fluid Reality, Chicago, IL, USA) and a vision-based tactile sensor (GelSight Inc., Waltham, MA, USA) into a custom 2-degree-of-freedom (DoF) telemanipulation system (Fig.~\ref{fig:overview}B) that also provides kinesthetic feedback via bilateral control. During telemanipulation, localized contact information is measured by the vision-based tactile sensor mounted on the robot finger and then mapped to binary actuation patterns on the tactile display positioned under the operator's fingertip (Fig.~\ref{fig:overview}B, C). The EEOP-based tactile display uses applied electric fields to move fluid through microchannels embedded in a printed circuit board (PCB), providing independent control of 32 tightly spaced tactile pixels (3\,mm hexagonal spacing; $13\ \mathrm{actuators/cm}^2$) and allowing us to render highly localized contact information to the operator. To our knowledge, this is the first teleoperation system that relays localized contact information from a robot fingertip to a human fingertip in this manner. To quantify behavioral naturalness, we compare teleoperated trajectories with those from direct human manipulation using dynamic time warping (DTW), which computes the deviation between two trajectories while accounting for temporal misalignment~\cite{sakoe}. Across two behavioral studies---a button-discrimination task and a peg-rolling task---we systematically vary the spatial resolution of the conveyed tactile signal, ranging from no feedback to highly localized feedback (Fig.~\ref{fig:overview}C), and find that highly localized tactile feedback produces teleoperated trajectories that more closely resemble natural human movement (Fig.~\ref{fig:overview}D). We further show that spatially distributed cutaneous feedback shrinks the state space covered by teleoperated demonstrations and improves trajectory consistency across operators with varying skill levels, reducing both within- and between-operator variability---properties that have been linked to improved learning performance in autonomous robot policies~\cite{mandlekar, belkhale}. Finally, to verify whether these improvements translate to autonomous manipulation, we fine-tune a $\pi_{0.5}$ policy~\cite{black} on demonstrations collected with and without spatially distributed tactile feedback. We find that the policy trained with spatially distributed tactile feedback achieves higher task success rates and exhibits faster, more humanlike behavior during autonomous execution. Together, these results suggest that preserving the spatial structure of tactile information alongside kinesthetic force feedback can make teleoperated demonstrations faster, more humanlike, and more consistent, enabling improved autonomous robot policies that more closely resemble human behavior.

\section*{RESULTS}

\begin{figure}
	\centering
	\includegraphics[width=0.8\textwidth]{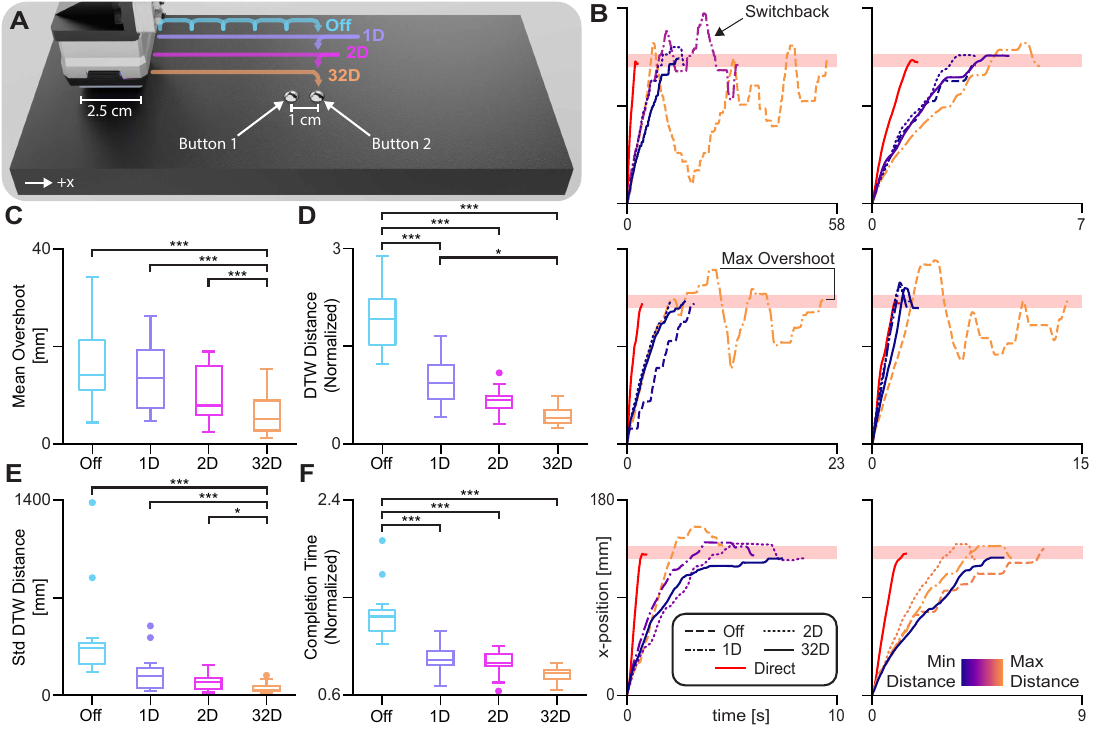}

    \singlespace
	\caption{\textbf{Button-discrimination task experimental results.}
        Box plots show median (center line), interquartile range (box), 1.5$\times$ interquartile range (whiskers), and outliers (circles). Unless otherwise noted, pairwise comparisons were performed between the \textit{32D} condition (orange) and the \textit{Off} (blue), \textit{1D} (purple), and \textit{2D} (magenta) conditions (stars indicate statistical significance).
		(\textbf{A}) Participants were asked to press one of two nearby buttons separated by 1\,cm without visual feedback. Idealized trajectories illustrate characteristic movement patterns observed under each feedback condition.
        (\textbf{B}) Trajectory of median distance from the participant's baseline trajectory (red) for each feedback condition (target: Button 2; odd-numbered participants shown). When participants received no cutaneous feedback, they frequently overshot the button and reversed directions (``switchbacks"). Higher resolutions of cutaneous feedback produced trajectories with a smaller DTW distance to the participant's baseline trajectory. Shaded region: button location.
        (\textbf{C}) Box plots of participants' mean overshoot distance for each condition.
        (\textbf{D}) Box plots of participants' mean DTW distance for each condition, normalized by their mean distance across all trials. Pairwise comparisons were performed between all six feedback-condition pairs.
        (\textbf{E}) Box plots of within-participant standard deviation of DTW distance.
        (\textbf{F}) Box plots of participants' mean completion time for each condition, normalized by their mean completion time across all trials. Pairwise comparisons were performed between all six feedback-condition pairs.
    }
	\label{fig:buttons}
\end{figure}

To evaluate the effects of high-resolution, spatially distributed tactile feedback on teleoperated manipulation, we conducted two experiments using a 2-DoF leader-follower telemanipulator. Spatially resolved contact patterns measured at the robot fingertip were reproduced via a 32-pixel tactile display at an update rate of 30\,Hz (see Materials and Methods). We systematically varied the spatial fidelity of the tactile feedback across four conditions (Fig.~\ref{fig:overview}C):

\begin{itemize}
\item \textbf{\textit{Off}:} No cutaneous feedback
\item \textbf{\textit{1D}:} All-in-phase cutaneous feedback
\item \textbf{\textit{2D}:} Two regions of cutaneous feedback
\item \textbf{\textit{32D}:} Fully localized cutaneous feedback
\end{itemize}

In both experiments, participants completed 48 trials (12 per feedback condition), and performance was quantified using both general and task-specific metrics. Participants also received kinesthetic force feedback across all cutaneous feedback conditions. Before each experiment, participants completed the task multiple times with their bare finger while it was strapped to the end of the follower device, allowing us to characterize natural task performance within the robot's kinematic workspace and compare the follower-device trajectories during teleoperation with the corresponding natural trajectories (see Supplementary Methods for specific details).

\subsection*{Spatially resolved feedback produces more natural teleoperated movements}

Humans routinely complete fine-motor tasks without direct visual attention---adjusting the volume on a smartphone, or pressing buttons on a television remote---by leveraging spatially varying feedback from across the fingerpad~\cite{polanen}. Reproducing this spatial structure in teleoperation should, in turn, allow operators to re-engage those same strategies and perform tasks as naturally as they would with their own hands. To evaluate this hypothesis, 12 participants were asked to push one of two adjacent buttons, spaced 1\,cm apart, without vision---a task that demands high operator precision, since imprecise finger positioning can depress both buttons at once (Fig.~\ref{fig:buttons}A). Across the 48 trials, a graphical user interface (GUI) instructed participants which of the two buttons to press, with the target button selected randomly on each trial. The participant was notified of trial completion only when the target button was pressed without accidentally pushing the adjacent button (see Supplementary Methods for more details).

Fully localized cutaneous feedback brought teleoperated behavior markedly closer to natural human performance. During direct task execution, participants followed largely ballistic trajectories toward the target button (Fig.~\ref{fig:buttons}B, red traces). Without cutaneous feedback, however, participants developed idiosyncratic strategies to complete the task, such as pressing randomly at various increments until the correct button was pushed (Fig.~\ref{fig:buttons}A, blue line). Coarser feedback helped, but often forced operators to overshoot and then reverse direction to isolate the target button without depressing its neighbor (Fig.~\ref{fig:buttons}A, purple and magenta lines). These detours resulted in teleoperated trajectories that exhibited larger overshoots (Fig.~\ref{fig:buttons}C) and more frequent direction reversals (``switchbacks"), sometimes several per trial (Fig.~\ref{fig:buttons}A, B). \textit{32D} feedback shifted both measures closer to the natural pattern, significantly reducing the average overshoot ($p < 0.001$; Table~\ref{tab:buttons_overshoot}) and switchback count ($p \leq 0.007$; Table~\ref{tab:buttons_switchbacks}) relative to the \textit{Off}, \textit{1D}, and \textit{2D} conditions.

To compare the direct and teleoperated movements more comprehensively, we computed the dynamic time warping (DTW) distance between each teleoperated trajectory and its direct-manipulation counterpart, with smaller DTW distances indicating closer agreement with natural movement patterns. DTW distance was strongly modulated by feedback resolution ($p < 0.001$; Fig.~\ref{fig:buttons}D); every feedback level (\textit{1D}, \textit{2D}, \textit{32D}) produced significantly smaller DTW distances than the \textit{Off} condition ($p < 0.001$), but improving naturalness beyond the \textit{1D} level required the high spatial resolution of the \textit{32D} condition ($p = 0.014$; Table~\ref{tab:buttons_dtw}). We then computed the standard deviation of each participant's DTW distances as a measure of within-participant variability. Notably, the variability of DTW distance in the \textit{32D} condition was significantly lower than in the \textit{2D} ($p = 0.013$), \textit{1D} ($p < 0.001$), and \textit{Off} ($p < 0.001$) conditions, indicating that high localization not only produced more natural trajectories, but did so more consistently (Fig.~\ref{fig:buttons}E; Table~\ref{tab:buttons_dtw_std}).

Completion time was likewise affected by feedback condition ($p < 0.001$; Fig.~\ref{fig:buttons}F): any level of cutaneous feedback accelerated task execution relative to the \textit{Off} condition ($p < 0.001$; Table~\ref{tab:buttons_time}). Unlike the DTW distance, however, the difference in completion times between the \textit{1D} and \textit{32D} conditions did not reach significance after correction for multiple comparisons ($p = 0.025,\ p_{\mathrm{adj}} = 0.074$), suggesting that while cutaneous feedback broadly accelerated task execution, higher feedback resolution yielded comparatively larger benefits for trajectory naturalness than for speed. Additionally, although the \textit{32D} condition did not yield significantly faster completion times than the \textit{1D} or \textit{2D} conditions, it exhibited significantly lower completion-time variability compared to the \textit{2D} ($p = 0.003$), \textit{1D} ($p = 0.003$), and \textit{Off} ($p < 0.001$) conditions (Table~\ref{tab:buttons_time_std}).

\subsection*{Spatially distributed feedback improves dynamic manipulation}

\begin{figure}
	\centering
	\includegraphics[width=0.8\textwidth]{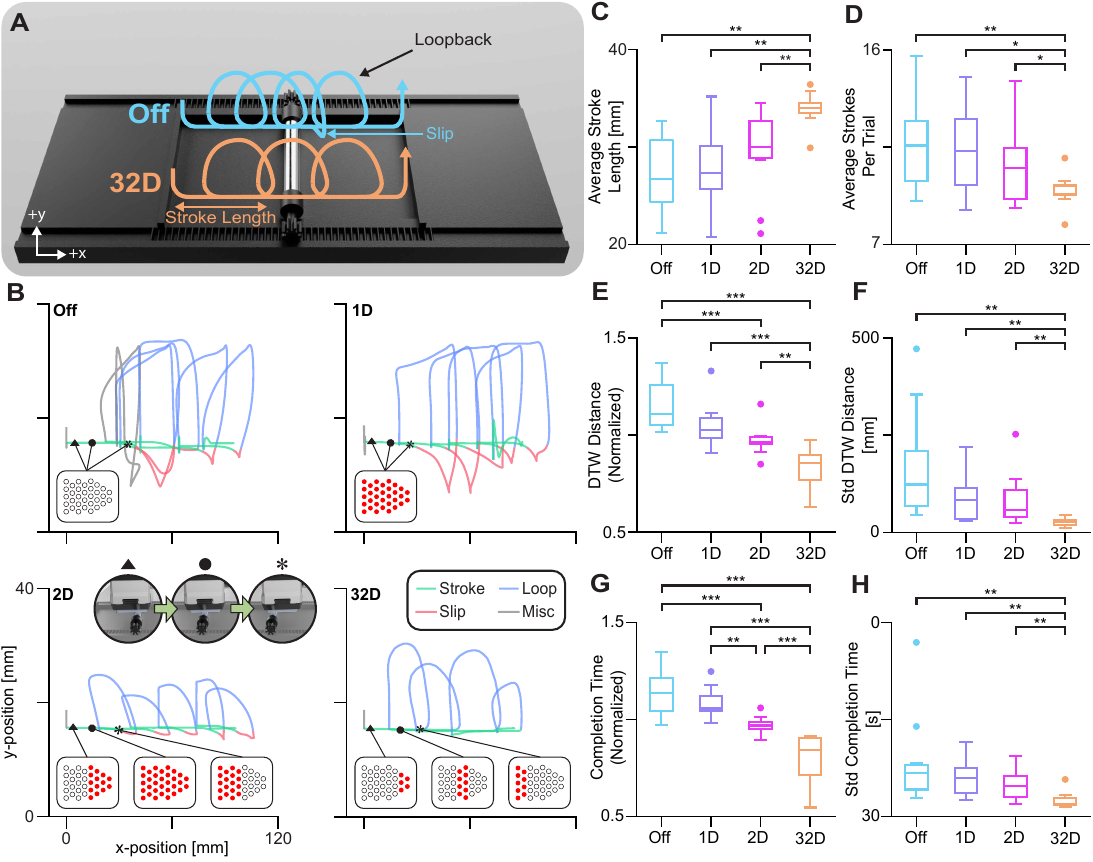}
    \singlespace
	\caption{\textbf{Peg-rolling task experimental results.}
        Box plots show median (center line), interquartile range (box), 1.5$\times$ interquartile range (whiskers), and outliers (circles). Unless otherwise noted, pairwise comparisons were performed between the \textit{32D} condition (orange) and the \textit{Off} (blue), \textit{1D} (purple), and \textit{2D} (magenta) conditions (stars indicate statistical significance).
        (\textbf{A}) Participants were asked to roll a peg forward to a hard stop and then roll it backward to its starting position. Task completion consisted of a series of rolling strokes punctuated by ``loopbacks" to reset the finger relative to the peg.
        (\textbf{B}) Prototypical trajectories from one participant during the forward-rolling phase, segmented into rolling strokes (green), loopbacks (blue), finger slips past the peg (red), and miscellaneous movements (gray). Example cutaneous-display activation patterns are shown at the beginning (triangle), middle (circle), and end (star) of a stroke.
        (\textbf{C}) Box plots of participants' mean stroke length for each condition.
        (\textbf{D}) Box plots of participants' mean number of strokes per trial for each condition.
        (\textbf{E}) Box plots of participants' mean DTW distance for each condition, normalized by their mean distance across all trials. Pairwise comparisons were performed between all six feedback-condition pairs.
        (\textbf{F}) Box plots of within-participant standard deviation of DTW distance.
        (\textbf{G}) Box plots of participants' mean completion time for each condition, normalized by their mean completion time across all trials. Pairwise comparisons were performed between all six feedback-condition pairs.
        (\textbf{H}) Box plots of within-participant standard deviation of completion time.
        }
	\label{fig:peg_rolling}
\end{figure}

In the button task, cutaneous feedback played a dominant role because kinesthetic cues reveal little about which button lies beneath the fingertip. To evaluate whether the benefits of spatially distributed feedback generalize to manipulation settings where cutaneous cues are less essential, 10 participants performed a visually occluded peg-rolling task in which they rolled a peg forward until it reached a hard stop and then rolled it backward to its starting position (Fig.~\ref{fig:peg_rolling}A). Unlike the button-discrimination task, this task does not involve multi-point tactile discrimination, and kinesthetic forces carry more task-relevant information by signaling whether the finger is in contact with the peg. As in the button task, participants completed 48 trials across 4 feedback conditions (12 per feedback condition).

Task completion consisted of a series of forward strokes punctuated by ``loopbacks" that reset the finger relative to the peg (Fig.~\ref{fig:peg_rolling}A, blue and orange lines). Optimal rolling therefore depended on a clear sense of where the finger sat relative to the peg, and the behavioral data suggest that full-resolution, \textit{32D} feedback supported this sense most effectively. This is evident in the prototypical trajectories taken from a single participant (Fig.~\ref{fig:peg_rolling}B). In the \textit{32D} condition, the participant completed the task in fewer strokes (Fig.~\ref{fig:peg_rolling}B, green lines); a clear sense of how far the peg had rolled across their finger allowed them to reposition their finger optimally (Fig.~\ref{fig:peg_rolling}B, blue lines) and maximize the length of the next stroke. In the \textit{2D} condition, the participant was less certain of the peg's location and consequently produced wasted motion by rolling off the peg (Fig.~\ref{fig:peg_rolling}B, red lines). In the \textit{1D} and \textit{Off} conditions, this uncertainty grew, and the participant tended to slip farther off the peg and onto the surface beneath. The \textit{Off} condition was the most severe: with no sense of the peg's position relative to the finger, the participant often looped back by the wrong amount and missed the peg entirely, producing spurious motions (Fig.~\ref{fig:peg_rolling}B, gray lines). Aggregated across all trials and participants, the \textit{32D} condition produced significantly fewer roll-offs onto the underlying surface ($p \leq 0.039$; Table~\ref{tab:peg_slips}), significantly longer strokes ($p \leq 0.004 $; Fig.~\ref{fig:peg_rolling}C; Table~\ref{tab:peg_stroke_length}), and significantly fewer strokes than any other condition ($ p \leq 0.014$; Fig.~\ref{fig:peg_rolling}D; Table~\ref{tab:peg_strokes}).

We then compared participants' teleoperated trajectories to their direct-manipulation counterparts. The differences in stroke length, number of strokes, and frequency of roll-offs were reflected in the DTW distances, which were strongly modulated by feedback resolution ($p < 0.001$; Fig.~\ref{fig:peg_rolling}E). Both \textit{2D} and \textit{32D} feedback yielded significantly smaller DTW distances (i.e., more natural trajectories) than the \textit{Off} condition ($p < 0.001$), and \textit{32D} feedback produced more natural trajectories than both the \textit{1D} ($p < 0.001$) and \textit{2D} ($p = 0.003$) conditions (Table~\ref{tab:peg_dtw}). Consistency followed the same pattern as in the button task: within-participant variability under \textit{32D} feedback was significantly lower than in any of the other conditions ($p \leq 0.005$; Fig.~\ref{fig:peg_rolling}F; Table~\ref{tab:peg_dtw_std}).

Completion time was also strongly affected by feedback resolution ($p < 0.001$; Fig.~\ref{fig:peg_rolling}G). Both \textit{2D} and \textit{32D} feedback produced faster completions than the \textit{Off} condition ($p < 0.001$); \textit{32D} feedback outpaced both \textit{1D} and \textit{2D} ($p < 0.001$); and \textit{2D} in turn was faster than \textit{1D} ($p = 0.001$; Table~\ref{tab:peg_time}). Completion times under \textit{32D} feedback were also more consistent than those under the coarser feedback conditions ($p = 0.003$; Fig.~\ref{fig:peg_rolling}H; Table~\ref{tab:peg_time_std}).

\subsection*{Improvements in naturalness co-occur with perceived reductions in mental demand}

\begin{figure}
	\centering
	\includegraphics[width=0.8\textwidth]{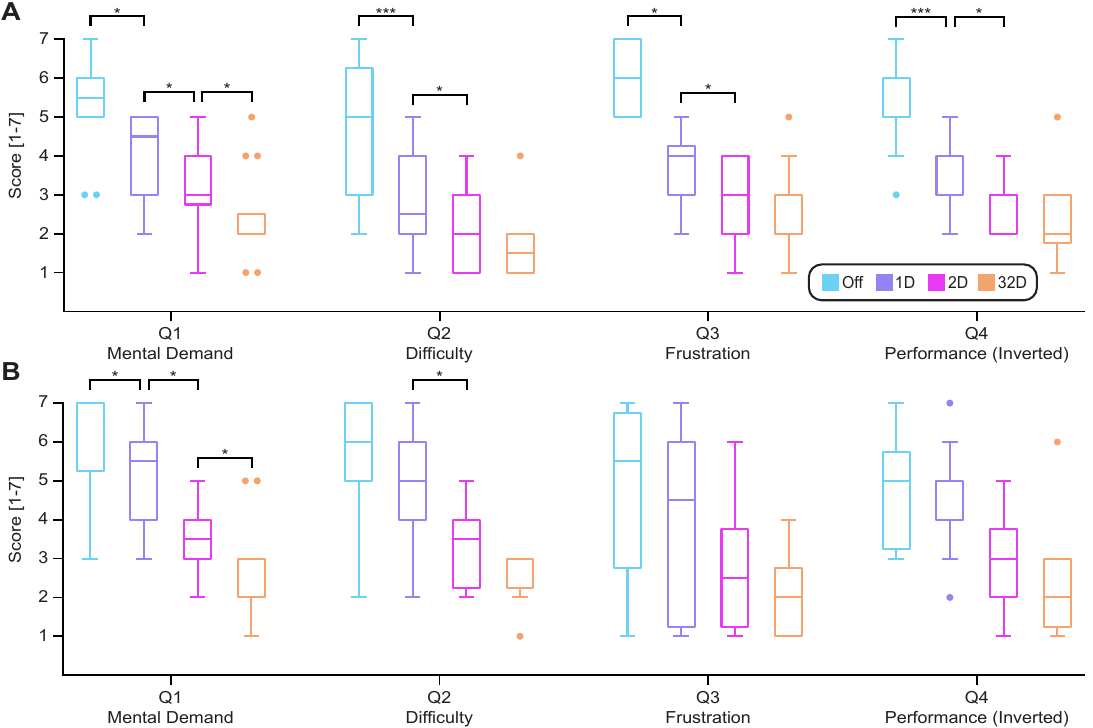}

	\caption{\textbf{Post-experiment questionnaire results.}
        After completing each task, participants completed a subset of the NASA-TLX questionnaire, rating perceived mental demand (Q1), task difficulty (Q2), frustration (Q3), and performance (Q4) on a 7-point scale. For Q1--Q3: 1 = low, 7 = high. Polarity of responses has been reversed for Q4: 1 = high, 7 = low. Box plots show median (center line), interquartile range (box), 1.5$\times$ interquartile range (whiskers), and outliers (circles). Stars indicate statistical significance.
		(\textbf{A}) Responses for the button-discrimination task.
        (\textbf{B}) Responses for the peg-rolling task.
    }
	\label{fig:questionnaire}
\end{figure}

The observed differences in movement naturalness and task completion time were accompanied by differences in participants' subjective experience across feedback conditions. In post-experiment questionnaires, participants rated the mental demand (Q1) and difficulty (Q2) of the task, their frustration during task execution (Q3), and their perceived performance (Q4) under each feedback condition. Notably, participants reported progressively lower mental demand with increasing feedback resolution (Fig.~\ref{fig:questionnaire}, Q1; Tables~\ref{tab:buttons_questionnaire} and \ref{tab:peg_questionnaire}). The co-occurrence of lower cognitive load with more natural, lower-variability trajectories suggests that high-resolution feedback may allow operators to engage the sensorimotor control strategies they already use in natural manipulation, rather than recruiting deliberate, cognitively and temporally costly strategies to interpret impoverished feedback signals.

\subsection*{Implications for learning from demonstration}

\begin{figure}
	\centering
	\includegraphics[width=0.8\textwidth]{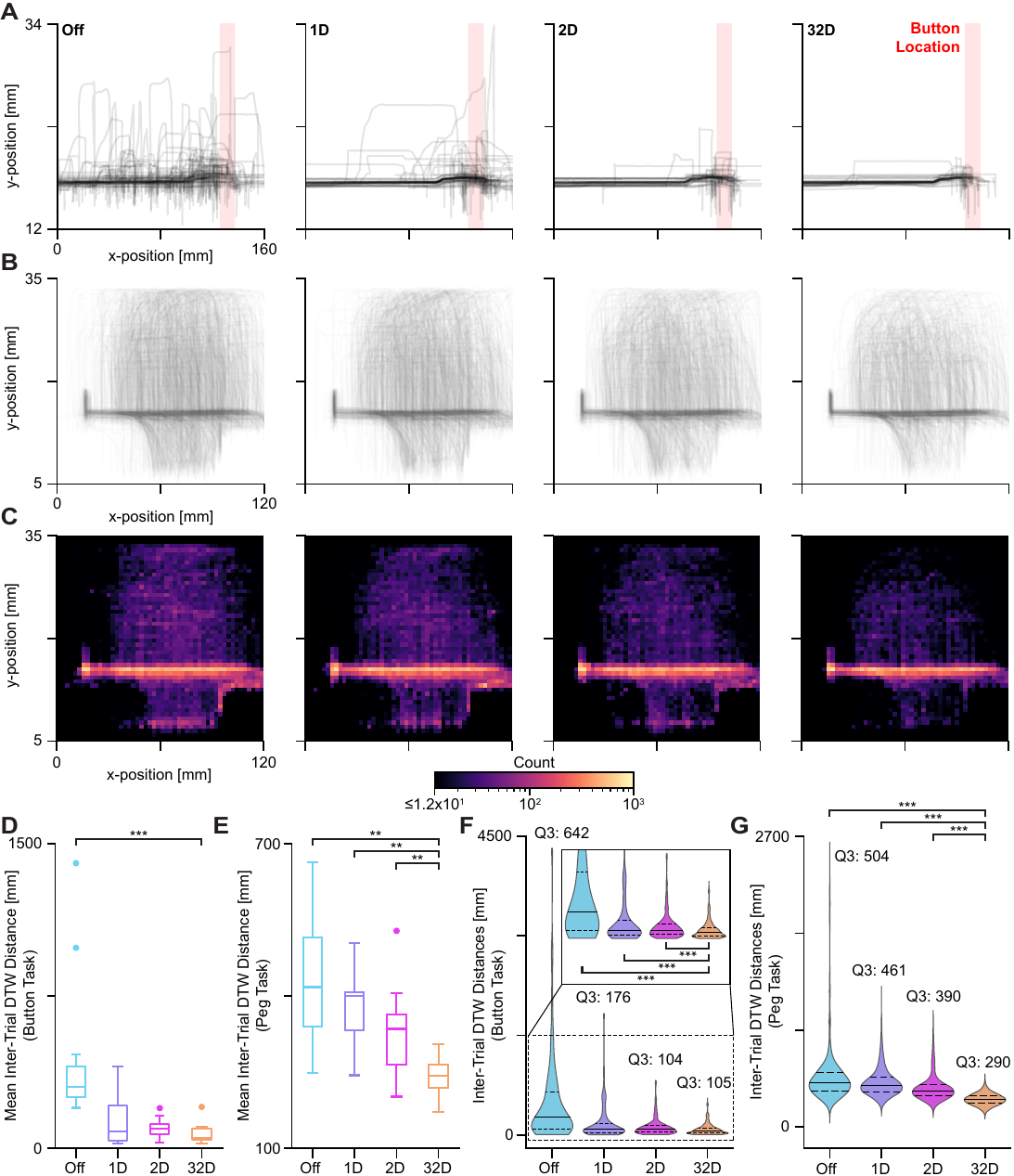}
    
    \singlespace
	\caption{\textbf{Improving dataset consistency for autonomous policy training.}
        Aggregating demonstrations collected with higher-resolution cutaneous feedback produces more concentrated trajectory distributions and reduced inter-trial variability.
		(\textbf{A}) Overlaid trajectories from all participants and trials in the button-discrimination task (target: Button 2) for each feedback condition.
        (\textbf{B}) Overlaid trajectories from all participants and trials in the peg-rolling task (forward roll only) for each feedback condition.
        (\textbf{C}) State-space occupancy heat maps for the peg-rolling task. Higher-resolution feedback yields a more concentrated state-space distribution.
        (\textbf{D}) Box plots of mean pairwise DTW distance between trials from the same participant in the button-discrimination task. Box plots show median (center line), interquartile range (box), 1.5$\times$ interquartile range (whiskers), and outliers (circles).
        (\textbf{E}) Box plots of mean pairwise DTW distance between trials from the same participant in the peg-rolling task. Box plot details: see above.
        (\textbf{F}) Violin plots of all pairwise inter-trial DTW distances in the button-discrimination task. Solid lines indicate medians and dashed lines indicate upper and lower quartiles. The upper quartile DTW distance is noted next to each of the respective plots.
        (\textbf{G}) Violin plots of all pairwise inter-trial DTW distances in the peg-rolling task. Violin plot details: see above.
    }
	\label{fig:dataset}
\end{figure}

Spatially distributed cutaneous feedback also improves the consistency and quality of datasets used for training autonomous robot policies. Prior work has shown that policy learning benefits from datasets whose trajectories occupy a smaller state space~\cite{belkhale}. We found that spatially distributed cutaneous feedback not only made teleoperation trials more natural, but also increased their consistency, resulting in a more compact state-space distribution. In the button task, spurious motions and incorrect button presses expanded the state-space coverage under the uniform or absent feedback conditions (Fig.~\ref{fig:dataset}A). In the peg-rolling task, the reduction in slips off the peg decreased occupancy of the lower region (y-position $< 15\,\mathrm{mm}$) of the state space (Fig.~\ref{fig:dataset}B). Because differences in state-space occupancy are difficult to discern from overlaid trajectories alone, we further visualized the trajectory distributions for the peg-rolling task using occupancy heat maps (Fig.~\ref{fig:dataset}C). These heat maps more clearly reveal that spatially distributed feedback produced more consistent demonstrations, concentrated within a smaller region of the state space.

To quantitatively assess trajectory consistency, we computed pairwise DTW distances between all trials within each feedback condition for every participant (Fig.~\ref{fig:dataset}D, E). These pairwise distances were then averaged within each participant to obtain a single measure of inter-trial variability per condition. Within participants, inter-trial DTW distance---a measure of trajectory variability---was significantly lower under spatially distributed cutaneous feedback compared to no cutaneous feedback, indicating more consistent demonstrations ($p \leq 0.003$; Tables~\ref{tab:buttons_mean_pairwise} and \ref{tab:peg_mean_pairwise}).

We then evaluated whether this effect extends to multi-operator datasets, where variability in teleoperation skill can increase dataset heterogeneity and degrade policy learning~\cite{mandlekar}. Participants in our studies reported a wide range of prior teleoperation experience, reflected by the variability in mean DTW distances and task completion times across participants (Fig.~\ref{fig:buttons}D, F; Fig.~\ref{fig:peg_rolling}E, G). To evaluate the effect of feedback on multi-operator datasets, we combined demonstrations from all participants to construct a separate dataset for each feedback condition (Fig.~\ref{fig:dataset}A, B) and computed DTW distances between all pairs of trials within each dataset. For both tasks, trajectory variability was significantly lower in the \textit{32D} dataset than in all other conditions ($p < 0.001$; Fig.~\ref{fig:dataset}F, G; Tables~\ref{tab:buttons_all_pairwise} and \ref{tab:peg_all_pairwise}), indicating that spatially distributed feedback yields more consistent datasets when demonstrations from multiple operators are aggregated.

Finally, to evaluate whether these improvements translate to better autonomous policy performance, we fine-tuned separate pre-trained $\pi_{0.5}$ policies~\cite{black} using demonstrations collected under the \textit{Off} and \textit{32D} conditions of the peg-rolling study (Fig.~\ref{fig:autonomy}A). Each policy was tested until 12 successful trials were obtained. The \textit{32D} policy completed the task more reliably, requiring 13 attempts to achieve 12 successes (92.3\% success) compared to the 19 attempts for the \textit{Off} policy (63.2\% success). To quantify the similarity of the autonomous behavior to natural human behavior, we compared each autonomous trial against the direct-manipulation trajectories captured in the peg-rolling study. The \textit{32D} policy exhibited significantly lower DTW distances ($p < 0.001$) and substantially lower variability than the \textit{Off} policy (Fig.~\ref{fig:autonomy}B, jitter plots), indicating that demonstrations collected with spatially distributed tactile feedback produced autonomous behaviors that were not only more humanlike but also more consistently humanlike. Additionally, the \textit{32D} policy achieved significantly faster task completion ($p < 0.001$) and lower completion-time variability than the \textit{Off} policy (Fig.~\ref{fig:autonomy}C, jitter plots).

\section*{DISCUSSION}

Across two tasks, we found that the spatial resolution of cutaneous feedback impacts not only how quickly operators complete a task, but also how closely their movements resemble direct human manipulation. Highly localized (\textit{32D}) cutaneous feedback drew teleoperated trajectories significantly closer to those produced during direct manipulation (Fig.~\ref{fig:buttons}B, D; Fig.~\ref{fig:peg_rolling}B, E) and did so more consistently across trials than coarser or absent feedback (Fig.~\ref{fig:buttons}E; Fig.~\ref{fig:peg_rolling}F). Whereas prior work has largely established that haptic feedback accelerates teleoperated task completion~\cite{pacchierotti, pacchierotti2016, meli, tavakoli}, our results show that the resolution of that feedback substantially influences the naturalness of the resulting motion.

\begin{figure}
	\centering
	\includegraphics[width=0.5\textwidth]{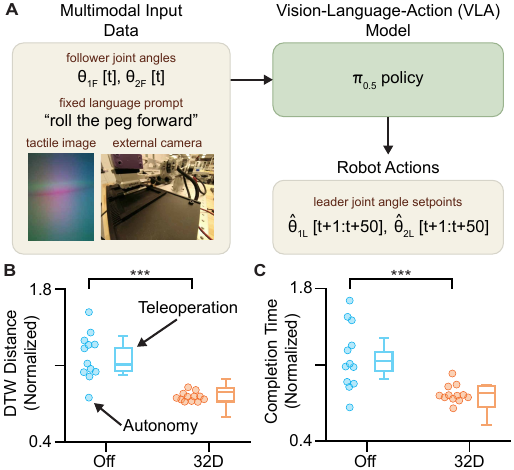}
    \singlespace
	\caption{\textbf{Autonomous peg-rolling policy results.}
        (\textbf{A}) A $\pi_{0.5}$ policy is fine-tuned to complete the forward-rolling phase of the peg-rolling task. The policy is conditioned on the follower joint angles ($\theta_{1\mathrm{F}}$ and $\theta_{2\mathrm{F}}$), an external camera image (right image), and a vision-based tactile sensor image (left image) at time $t$, and it generates a 50-step chunk of leader positions ($\hat{\theta}_{1\mathrm{L}}$ and $\hat{\theta}_{2\mathrm{L}}$) for time $t+1$ to $t+50$.
        (\textbf{B}) Jitter plots of normalized DTW distances for each autonomous trajectory generated by the \textit{Off} (blue) and \textit{32D} (orange) policies. Box plots of normalized DTW distances for teleoperated peg-rolling trials are shown to the right of each jitter plot for reference. Box plots show median (center line), interquartile range (box), 1.5$\times$ interquartile range (whiskers), and outliers (circles). Stars indicate statistical significance.
        (\textbf{C}) Jitter plots of normalized completion times for each autonomous trajectory generated by the \textit{Off} and \textit{32D} policies. Box plots of normalized completion times for teleoperated peg-rolling trials are shown to the right of each jitter plot for reference. Box plot details: same as above.
    }
	\label{fig:autonomy}
\end{figure}

Additionally, spatially distributed feedback improves the quality and consistency of datasets used to train autonomous robot policies. Prior studies have shown that policy learning can be hindered by demonstrations exhibiting excessive variability, diffuse state-space coverage, or large differences in operator proficiency~\cite{mandlekar, belkhale}. In line with these findings, we found that spatially distributed feedback produced trajectories with more concentrated state-space occupancy (Fig.~\ref{fig:dataset}A, B, C) and greater consistency both within (Fig.~\ref{fig:dataset}D, E) and across (Fig.~\ref{fig:dataset}F, G) operators.

Finally, we demonstrated that these improvements in teleoperated performance carry over to autonomous policies trained on the resulting demonstrations, leading to higher success rates, faster task completion, and more humanlike behavior during autonomous execution. In the peg-rolling task, the \textit{32D} feedback reduced teleoperated task completion time by an average of 33\% and deviation from natural human behavior by 28\% compared to the \textit{Off} condition. The resulting \textit{32D} policy exhibited comparable benefits, completing the task 29\% faster, on average, and with 31\% less deviation from human behavior than the policy trained on demonstrations without tactile feedback (Fig.~\ref{fig:autonomy}B, C). These findings demonstrate that improving the quality of teleoperated demonstrations through spatially distributed tactile feedback can enhance both the efficiency and human-likeness of learned autonomous behavior.

\subsection*{The value of feedback resolution depends on the task}

\subsubsection*{Kinesthetic interactions}

Although spatially distributed cutaneous feedback consistently improved performance across both tasks, the magnitude and nature of the benefit depended on the task itself. One task-dependent factor was the degree of redundancy between kinesthetic and cutaneous information, as uniform cutaneous feedback becomes redundant with kinesthetic force cues when the two are aligned. In the peg-rolling task, many participants reported that the uniform \textit{1D} feedback felt very similar to the \textit{Off} condition. We attribute this to redundancy with kinesthetic cues: pushing down on the peg transmits forces through the telemanipulator that already deform the operator's fingerpad uniformly, supplying much of what the spatially uniform \textit{1D} rendering would convey. Consistent with this interpretation, the reductions in DTW distance ($p = 0.059$) and completion time from \textit{Off} to \textit{1D} ($p = 0.112$) were both nonsignificant. In the button-discrimination task, by contrast, kinesthetic cues carried little information about button location, leaving the cutaneous channel as the dominant source of task-relevant information; there, even uniform \textit{1D} feedback produced large and highly significant improvements over \textit{Off} in both naturalness ($p < 0.001$) and speed ($p < 0.001$).

The redundancy of kinesthetic and uniform cutaneous cues motivates sensory subtraction approaches, in which spatially uniform cutaneous deformation is used to convey force information in place of kinesthetic feedback~\cite{pacchierotti, pacchierotti2016, meli}. Our results add an important qualification: when tactile feedback provides \textit{spatially distributed} information about the contact, it provides information that is distinct from and complementary to kinesthetic feedback. The added spatial information can accelerate task completion and yield trajectories that are significantly more natural and consistent.

\subsubsection*{Feature size}
In some cases, coarser feedback levels can help amplify small features. In the button task, although many participants found \textit{32D} feedback the most intuitive, some reported that \textit{2D} was easier to interpret: under \textit{32D} feedback, each small button inflated only a few pixels on the cutaneous display, whereas the coarser \textit{2D} rendering expanded the size of each button to half the display. Accordingly, \textit{32D} feedback provided little additional benefit over \textit{2D} (naturalness, $p = 0.244$; speed, $p = 0.237$). In the peg-rolling task, however, the peg inflated one to two rows of pixels at a time, producing larger and more perceivable contact patches---mirrored by the significantly more natural and faster trajectories under \textit{32D} than \textit{2D} reported above. Taken together, these observations suggest that an intermediate resolution could have offered a better trade-off between spatial detail and feature salience in the button-discrimination task. For example, an \textit{8D} condition, in which each row of tactile pixels is treated as a separate zone, might have made the buttons easier to localize than the \textit{2D} condition while being easier to perceive than the \textit{32D} condition.

These observations regarding feature size, in conjunction with the above observations about kinesthetic interactions, suggest that the value of cutaneous feedback resolution is not fixed but depends on the task's spatial information requirements. Highly localized tactile feedback is most valuable when it conveys information that is unavailable through kinesthetic feedback and when the relevant contact features can be effectively represented by the display. Thus, rather than treating increased resolution as uniformly beneficial, teleoperation systems should match the tactile rendering resolution to the spatial-information demands of the task.

\subsection*{Improving proprioceptive awareness}

The peg-rolling task further hints at a broader role for cutaneous feedback in supporting participants' proprioceptive sense of where the manipulator is positioned with respect to the environment. Although peg rolling does not require fine tactile discrimination in the same way the button task does, participants frequently reported that the \textit{32D} condition gave them a clearer sense of how far the peg had traveled across the fingerpad, allowing them to better judge how far to reposition the finger during each loopback. Behaviorally, this manifested in longer stroke lengths and fewer strokes per trial, consistent with operators using the spatial cutaneous signal to plan their repositioning. Because cutaneous and proprioceptive afferents are known to be integrated during natural sensorimotor control~\cite{collins, proske}, it is plausible that richer cutaneous rendering sharpens the state estimate used to plan movement. However, additional studies are required to establish a causal link between rendered cutaneous resolution and proprioceptive awareness.

\subsection*{The case for naturalness}

Our results suggest that naturalness and task speed, while strongly associated, are separable performance dimensions that can improve independently. In the button-discrimination task, spatially distributed tactile feedback reduced task completion time by an average of 46\%, while decreasing deviation from natural human behavior by 77\%. Thus, even when improvements in task speed are relatively modest, spatially distributed contact information can yield further meaningful improvements by producing more natural demonstrations with other desirable qualities. For example, in the button-discrimination task, more natural demonstrations exhibited smaller overshoots (Fig.~\ref{fig:buttons}C) and fewer direction reversals (Table~\ref{tab:buttons_switchbacks}). Similarly, in the peg-rolling task, more natural demonstrations were characterized by longer rolling strokes (Fig.~\ref{fig:peg_rolling}C), fewer strokes required to complete the task (Fig.~\ref{fig:peg_rolling}D), and fewer slips off the peg. Together, these behaviors produced more concentrated state-space coverage in the \textit{32D} condition than in the coarser feedback conditions (Fig.~\ref{fig:dataset}A, B, C). Moreover, completion-time variability was significantly lower when participants received spatially distributed feedback (Table~\ref{tab:buttons_time_std}), even when task completion times were not significantly reduced (Table~\ref{tab:buttons_time}), indicating greater consistency across demonstrations. The reduced variability in both the spatial and temporal characteristics of teleoperated demonstrations was reflected in the autonomous policy trained on \textit{32D} demonstrations, which achieved higher task success rates, faster autonomous execution with lower completion-time variability, and more humanlike manipulation behaviors with greater consistency across autonomous rollouts.

Variability, of course, has benefits: it has been shown that diversity across environments, initial conditions, task instances, and successful manipulation strategies can improve the robustness and generalization of learned policies~\cite{yue, walke}. However, this is not an argument for low-quality teleoperation, as not all forms of variability contribute equally to learning~\cite{belkhale, mandlekar}. For example, prior work has shown that haptic feedback can reduce both the magnitude and the variability of forces applied during demonstrations, yielding autonomous policies that complete tasks with higher success rates while applying lower and more consistent forces~\cite{tan}. In such cases, variability in applied forces reflects an inconsistency in execution rather than useful information about failure recovery or alternative manipulation strategies. Similarly, our results demonstrate that spatially distributed tactile feedback primarily reduced erroneous movements, such as overshoots, rather than limiting meaningful behavioral diversity. As a result, demonstrations collected with spatially distributed feedback can provide a cleaner supervisory signal for imitation learning while preserving opportunities to capture the diversity of valid manipulation strategies within a task itself. This improvement may also complement a number of manual~\cite{friedrich} and statistical~\cite{chuck, krishnan} data curation methods, which identify and filter noisy and low-quality demonstrations from training datasets. By reducing execution-related variability at the point of data collection, spatially distributed feedback may allow a greater proportion of demonstrations to be retained during curation, potentially reducing the number of demonstrations that must be collected to assemble a high-quality dataset~\cite{cuan}.

Beyond implications for autonomous policy learning, the distinction between task speed and naturalness may also be important for applications in which operators perform repetitive manipulations over long durations, where reducing cognitive effort and promoting intuitive control could be as valuable as maximizing throughput. Because higher resolutions of tactile feedback reduced participants' self-reported mental demand (Fig.~\ref{fig:questionnaire}, Q1), spatially distributed cutaneous feedback may support longer, less fatiguing teleoperation sessions, though dedicated studies will be needed to evaluate operator endurance directly.

\subsection*{Limitations and future directions}

Several limitations should be considered when interpreting these results. First, the present study employed a 2-DoF leader-follower telemanipulator and two experimental tasks chosen to span distinct manipulation demands. Although these tasks reveal consistent benefits of spatially resolved cutaneous feedback, they do not encompass the full range of contact interactions encountered in real-world teleoperation. Future studies should investigate whether the observed benefits extend to higher-DoF systems, multi-finger telemanipulation, and more complex dexterous tasks. Furthermore, determining the spatial resolution of tactile feedback required to reproduce natural manipulation across different tasks, and identifying where additional resolution yields diminishing returns, would provide valuable design targets for future tactile displays and teleoperation systems. Similarly, the level of control required over individual tactile elements is an important consideration. The tactile displays used in this study supported only binary pixel actuation, and although future displays may enable control of indentation depth, the extent to which variable-depth actuation improves performance relative to binary actuation remains to be established.

Additionally, this study evaluated spatially distributed tactile feedback only in conjunction with kinesthetic force feedback. We hypothesize that many of the observed benefits would persist in the absence of kinesthetic feedback, and integrating spatially distributed tactile feedback into ungrounded motion-tracking gloves may still improve teleoperation performance. This hypothesis is consistent with prior work demonstrating that tactile feedback can convey some of the information otherwise provided by kinesthetic force feedback \cite{pacchierotti, pacchierotti2016, meli}, but further experiments are required to establish its validity. Finally, the potential benefits of incorporating additional feedback modalities, such as vibrotactile feedback, also warrant investigation. Although prior work suggests that low-frequency tactile feedback, as provided by the displays used in this study, accounts for most of the performance improvements observed when combined with high-frequency feedback~\cite{wildenbeest}, vibrotactile feedback may nevertheless provide complementary information, including cues about friction and slip, that further supports dexterous manipulation~\cite{johansson, johansson_signals, chen}.

\subsection*{Conclusion}
 
Reproducing the spatial structure of contact experienced by a remote manipulator is important for natural teleoperation and could improve performance across applications ranging from hazardous-environment manipulation to robot learning from demonstration. Across two manipulation tasks, we found that augmenting kinesthetic force feedback with spatially distributed cutaneous feedback substantially narrowed the gap between teleoperated movements and direct manipulation, improving not only task speed, but also the naturalness, consistency, and perceived cognitive demands of teleoperation. Beyond these immediate performance gains, our results suggest that spatially distributed cutaneous feedback may strengthen an operator's proprioceptive awareness, helping them better estimate contact state and plan subsequent movements. More broadly, as teleoperation increasingly serves as a primary mechanism for collecting demonstration data to train autonomous manipulation systems~\cite{argall, jaquier, zhao, chi}, we show that the quality of tactile feedback influences not only the human operator's performance but also the quality of policies learned from their demonstrations. By promoting more natural and consistent behavior, spatially resolved cutaneous feedback improves both human teleoperation and the autonomous systems that learn from it.

\section*{MATERIALS AND METHODS}

\subsection*{Experimental hardware}

Experimentation was performed on a custom 2-DoF leader-follower teleoperation system capable of horizontal and vertical motion (Fig.~\ref{fig:overview}B). The horizontal and vertical DoFs are each actuated by a brushless DC motor (EC 45 Flat, Maxon Group, Sachseln, Switzerland) through cable transmissions with 33:1 and 30:1 reduction ratios, respectively, while link positions are measured by 14-bit capacitive encoders (AMT212B-V, Same Sky, Lake Oswego, OR, USA). Each DoF is controlled by a separate microcontroller (Teensy 4.1, SparkFun Electronics, Boulder, CO, USA) running field-oriented commutation through the SimpleFOC library, with resulting signals routed to four motor drivers (BOOSTXL-DRV8301, Texas Instruments, Dallas, TX, USA) that power the motors. A bilateral controller is executed at 10 kHz on each microcontroller to deliver kinesthetic force feedback to the human operator. The robot ``finger" is a vision-based tactile sensor (GelSight Mini, GelSight Inc., Waltham, MA, USA), and cutaneous feedback is provided to the human operator using a 32-pixel cutaneous shape display driven by embedded electro-osmotic pumps~\cite{shultz, shen} (Fluid Reality, Chicago, IL, USA; Fig.~\ref{fig:overview}C). The vision-based tactile sensor captures video at 30\, frames per second (fps), and the video frames are converted to binary actuation patterns on the tactile display at 30\,Hz. The pixels on the tactile display are arranged in a hexagonal pattern with a spacing of 3\,mm, with each pixel having a diameter of 1.8\,mm and a maximum no-load displacement of approximately 1\,mm (Fig.~\ref{fig:overview}C). When actuated, the pixels reach 80\% of their maximum displacement within 200\,ms and return from full displacement to 0\% displacement within 150\,ms.

\subsection*{Mapping from tactile sensor to shape display}

When the robot finger makes contact with the environment (Fig.~\ref{fig:mapping}A), $240~\times~320$-pixel images from the vision-based tactile sensor (Fig.~\ref{fig:mapping}B) are converted into a 32-pixel inflation pattern (Fig.~\ref{fig:mapping}C) corresponding to the physical layout of the shape display (Fig.~\ref{fig:mapping}D; Algorithm~\ref{alg:mapping}). When the tactile sensor is first initialized, a reference image $I_\mathrm{0}$ is captured and stored. Each subsequent frame $I$ (Fig. \ref{fig:mapping}B) is then differenced with $I_\mathrm{0}$ and converted to grayscale to yield a difference image $I_\mathrm{d} = I - I_\mathrm{0}$, which is smoothed with a $3 \times 3$ Gaussian blur. If the Frobenius norm of $I_\mathrm{d}$ falls below a certain threshold, $T_{\mathrm{noise\_floor}}$, no pixels are inflated on the shape display. Otherwise, Otsu's method is applied to binarize $I_\mathrm{d}$ via adaptive histogram-based thresholding. Each pixel $i$ on the cutaneous display is then mapped to a corresponding image coordinate $(x_i, y_i)$, reflecting the physical layout of the display (visualized as circles of radius $r$ in Fig.~\ref{fig:mapping}C). If the Frobenius norm of pixel intensities within a $2r \times 2r$ region centered at each $(x_i, y_i)$ exceeds a contact threshold $T_\mathrm{contact}$ (shaded circles in Fig.~\ref{fig:mapping}C), the corresponding display pixel is inflated to convey localized contact to the operator's fingertip (Fig.~\ref{fig:mapping}D). Additional mapping details can be found in the Supplementary Methods.

\subsection*{Experimental design}

We evaluated the effect of cutaneous feedback resolution on task naturalness using two tasks: a button-discrimination task and a peg-rolling task. All participants were na\"{i}ve to the goals and design of the study, consented to participate in the experiment under Northwestern University's Institutional Review Board protocol \#STU00223620, and were compensated for their time. In both tasks, participants performed 48 trials across 4 feedback conditions (12 trials per condition, randomized order; Fig.~\ref{fig:overview}C):

\begin{itemize}
\item \textbf{\textit{Off}:} No cutaneous feedback
\item \textbf{\textit{1D}:} All-in-phase cutaneous feedback
\item \textbf{\textit{2D}:} Two regions of cutaneous feedback
\item \textbf{\textit{32D}:} Fully localized cutaneous feedback
\end{itemize}

In the \textit{Off} condition, no shape display pixels were inflated regardless of contact on the robot finger. In the \textit{1D} condition, all pixels were inflated if any single pixel was activated by the mapping algorithm (Algorithm~\ref{alg:mapping}). In the \textit{2D} condition, all pixels in the top or bottom half of the display were inflated if any pixel in the corresponding half was activated. Finally, in the \textit{32D} condition, only the pixels activated by the mapping algorithm were inflated. Participants received kinesthetic force feedback in all 48 trials, and the robot end-effector trajectory was recorded throughout each trial.

Before the teleoperation trials, each participant's dominant index finger was strapped to the follower robot's end-effector, and the participant completed the task directly with their bare finger. Participants were blindfolded to match the absence of visual feedback during teleoperation, and the resulting trajectories served as baselines for comparison (see Fig.~\ref{fig:buttons}B, red traces). During the actual trials, participants' view of the task workspace and robot end-effector was obstructed by a monitor displaying a graphical user interface (GUI). The GUI conveyed the current trial number, the active cutaneous feedback condition, and task-specific instructions. Following the trials, participants responded to the following selection of questions from the NASA Task Load Index (TLX) questionnaire~\cite{hart} using a scale of 1 (Not at all) to 7 (Very):

\begin{itemize}

\item \textbf{Mental Demand (Q1):} How mentally demanding was the task?
\item \textbf{Difficulty (Q2):} How hard did you have to work to accomplish your level of performance?
\item \textbf{Frustration (Q3):} How insecure, discouraged, irritated, stressed, and annoyed were you?
\item \textbf{Performance (Q4):} How successful were you in accomplishing what you were asked to do?
\end{itemize}

\subsubsection*{Button discrimination}
Twelve participants between the ages of 19 and 37 (7 male, 5 female, mean age 28.6 years) took part in this experiment, of whom 11 were right-handed and 1 was left-handed. No participant reported physical impairments that might have affected their ability to operate the device or perceive the haptic feedback. The participants had varying levels of experience with teleoperation devices: 5 reported no experience, 6 reported some experience, and 1 reported having extensive experience. Participants were instructed to find and press one of two nearby buttons spaced 1\,cm apart---close enough that the robot finger (2.5\,cm across) could simultaneously cover and depress both (Fig.~\ref{fig:buttons}A). The buttons were constructed using 6.35-mm-diameter ball bearings, enabling the tactile sensor to slide over them without damaging the gel of the sensor (see Fig.~\ref{fig:ball_bearing_button} for additional details). The buttons were electrically connected to a data acquisition program and required a 0.25-mm displacement with a peak applied force of 2.7\,N to register a click. In each trial, the GUI instructed the participant to push either the closer or farther button; the trial concluded only when the correct button (and no other button) was pressed. Between trials, the button positions were varied between two possible locations along the forward--backward direction of motion to prevent participants from memorizing their locations (Fig.~\ref{fig:task_layout}, left panel). Participants were given 8 practice trials (2 per feedback condition) before the start of the actual trials.

\subsubsection*{Peg rolling} A separate set of 10 participants between the ages of 19 and 31 (7 male, 3 female, mean age 25.9 years) took part in the peg-rolling experiment, all of whom were right-handed. No participant reported physical impairments that might have affected their ability to operate the device or perceive the haptic feedback. The participants had varying levels of experience with teleoperation devices: 4 reported no experience, 6 reported some experience, and 0 reported having extensive experience. Participants were instructed to roll a peg forward until it contacted a hard stop, and then roll it backward to its starting position. The peg was constructed out of a steel shaft and, to ensure that the peg remained straight throughout each trial, it was specially designed with a rack and pinion on both sides (Fig.~\ref{fig:peg_rolling}A). The GUI directed participants when to initiate each movement phase (i.e., ``Roll Forward" and ``Roll Backward"), and an external camera recorded task execution during the trials. Participants were given 4 practice trials (1 per feedback condition) before beginning the actual trials.

\subsection*{Data and statistical analysis}

During all trials, trajectory data were collected at 50 Hz. For both tasks, two-dimensional dynamic time warping (DTW) distances were computed between each teleoperated trajectory and each of the participant's equivalent direct-manipulation trajectories, then averaged to produce a single DTW distance per trial. Distances were then normalized by each participant's mean DTW distance across all trials. A linear mixed-effects model (LME) with random intercepts was fit to the normalized DTW distances, with Feedback Condition as a fixed effect and Participant as a random effect:

\vspace*{-\baselineskip}

\begin{equation}
	\mathrm{Normalized\ DTW\ Distance} \sim \mathrm{Feedback\ Condition} + (1 | \mathrm{Participant}).
	\label{eq:dtw_model}
\end{equation}

\vspace{-4pt}

\noindent The main effect of Feedback Condition was evaluated using analysis of variance (ANOVA; \mbox{$\alpha = 0.05$}), followed by post hoc coefficient tests between all six pairs of conditions ($\alpha = 0.05$, Holm-Bonferroni corrected). To assess variability in task naturalness, per-participant standard deviations of DTW distance were compared between the \textit{32D} condition and the \textit{2D}, \textit{1D}, and \textit{Off} conditions using one-tailed Wilcoxon signed-rank tests ($\alpha = 0.05$, Holm-Bonferroni corrected).

Task completion times were normalized by each participant's mean completion time across all trials and then analyzed using the same LME structure:

\vspace*{-\baselineskip}

\begin{equation}
	\mathrm{Normalized\ Completion\ Time} \sim \mathrm{Feedback\ Condition} + (1 | \mathrm{Participant}).
	\label{eq:time_model}
\end{equation}

\vspace{-4pt}

\noindent The main effect of Feedback Condition was evaluated with an ANOVA ($\alpha = 0.05$), followed by post hoc pairwise coefficient tests ($\alpha = 0.05$, Holm-Bonferroni corrected) between all conditions. To assess variability in task completion times, per-participant standard deviations of completion time were compared between the \textit{32D} condition and the \textit{2D}, \textit{1D}, and \textit{Off} conditions using one-tailed Wilcoxon signed-rank tests ($\alpha = 0.05$, Holm-Bonferroni corrected).

Task-specific trajectory metrics were also analyzed. For the button task, the mean overshoot beyond the target button and mean number of switchbacks were compared. For the peg-rolling task, the mean stroke length, mean number of strokes per trial, and mean number of slips from the peg onto the underlying surface were compared. For each metric, one-tailed Wilcoxon signed-rank tests ($\alpha = 0.05$, Holm-Bonferroni corrected) were performed between the \textit{32D} condition and the \textit{2D}, \textit{1D}, and \textit{Off} conditions.

To quantify the consistency of teleoperated trajectories, we computed pairwise DTW distances between all teleoperated trials within each feedback condition. We first computed pairwise DTW distances between trajectories from the same participant, yielding a mean inter-trial DTW distance for each participant and feedback condition. These per-participant means were compared between the \textit{32D} condition and the \textit{2D}, \textit{1D}, and \textit{Off} conditions using one-tailed Wilcoxon signed-rank tests ($\alpha = 0.05$, Holm-Bonferroni corrected). We then combined trajectories across participants to form an aggregated demonstration dataset for each feedback condition, mimicking the multi-operator datasets often used for robot learning. Pairwise DTW distances were computed between all trajectories in each aggregated dataset to characterize inter-trial variability across operators. The resulting DTW-distance distributions were compared between the \textit{32D} condition and each of the other feedback conditions using one-tailed Mann-Whitney \textit{U} tests ($\alpha = 0.05$, Holm-Bonferroni corrected).

Finally, post-experiment questionnaire responses (Q1--Q4) were analyzed using one-tailed Wilcoxon signed-rank tests ($\alpha = 0.05$, Holm-Bonferroni corrected). Comparisons were performed between adjacent feedback conditions, i.e., between the \textit{Off} and \textit{1D}, \textit{1D} and \textit{2D}, and \textit{2D} and \textit{32D} conditions.

The statistical results for all the tests mentioned above can be found in the Supplementary Text and Tables S1 to S19.

\subsection*{Autonomous policy learning}

To evaluate the effects of demonstration quality on autonomous robot behavior, we fine-tuned separate instances of a pre-trained $\pi_{0.5}$ policy~\cite{black} using demonstrations collected under the \textit{Off} and \textit{32D} feedback conditions of the peg-rolling study. The policies were trained to perform forward peg rolling; accordingly, the training datasets included only the forward-rolling phase of participants' demonstrations. The policies were conditioned solely on current observations to predict future actions. Each policy received images from the vision-based tactile sensor and external camera, follower joint positions, and a fixed language prompt (``roll the peg forward") as input, and generated a 50-step chunk of leader-position setpoints (Fig.~\ref{fig:autonomy}A). All model parameters were fine-tuned for 3000 steps using a batch size of 32, the AdamW optimizer, and a peak learning rate of $2.5\times10^{-5}$ with cosine decay. Fine-tuning was performed on an RTX6000 GPU (NVIDIA Corporation, Santa Clara, CA, USA). Additional details about policy training and autonomous rollouts are provided in the Supplementary Methods.

We executed each autonomous policy until 12 successful trials had been collected per policy. For each successful autonomous trajectory, we calculated its DTW distance to each of the 60 direct human manipulation trajectories collected during the peg-rolling study and averaged these distances to obtain a single measure of behavioral similarity per autonomous trial. To maintain consistency with the preceding analyses, all DTW distances and completion times were normalized by the mean over all 24 successful autonomous trials prior to statistical analysis. Finally, one-tailed Mann-Whitney \textit{U} tests ($\alpha = 0.05$) were performed to compare the DTW distances and completion times between the autonomous policies trained on \textit{Off} and \textit{32D} demonstrations. The results of these tests are provided in the Supplementary Text.

\section*{ACKNOWLEDGMENTS}
The authors thank Matthew Elwin, Carl Moore, and Rodney Roberts for their guidance on this project. The authors also thank Yufeng Yang, Andrew Pavlovic, and Alyssa Chen for their contributions to the leader-follower device, and Tony Shilati and Charles De Mey for setting up instrumentation used to characterize several devices in this work. Finally, the authors thank Joe Mullenbach, Craig Shultz, Gabriel Beutel, Joshua Jung, and German Espinosa of Fluid Reality for providing the tactile displays that were essential to these experiments.

\subsection*{Funding}
This material is based upon work supported by the National Science Foundation under Grant NRI-2221571 and Grant 2330040, the HAND Engineering Research Center (ERC). Rohan Kota was supported by the Department of Defense (DoD) through the National Defense Science \& Engineering Graduate (NDSEG) Fellowship Program.

\subsection*{Author contributions}
RK, GR, and JEC conceived the study and formulated the central research idea. RK built the experimental hardware. RK and GR designed the experiments. RK conducted the participant studies and performed the data analysis. RK, GR, and JEC interpreted the results and wrote and edited the manuscript.

\subsection*{Competing interests}
The authors report no competing interests.

\subsection*{Data, code and materials availability}
All data needed to support the conclusions of this work can be found in the main text or Supplementary Materials.

\section*{SUPPLEMENTARY MATERIALS}
Supplementary Methods\\
Supplementary Text\\
Algorithm S1 \\
Figs. S1 to S4\\
Tables S1 to S19 \\

\clearpage 


\newpage


\renewcommand{\thefigure}{S\arabic{figure}}
\renewcommand{\thetable}{S\arabic{table}}
\renewcommand{\theequation}{S\arabic{equation}}
\renewcommand{\thealgorithm}{S\arabic{algorithm}}
\renewcommand{\thepage}{S\arabic{page}}
\setcounter{figure}{0}
\setcounter{table}{0}
\setcounter{equation}{0}
\setcounter{page}{1} 


\clearpage
\setcounter{page}{1}
\renewcommand{\thepage}{S\arabic{page}}

\title{\bfseries\boldmath Supplementary Materials for \\ \scititle}
\maketitle

\subsection*{This PDF file includes:}
Supplementary Methods\\
Supplementary Text\\
Algorithm S1\\
Figures S1 to S4\\
Tables S1 to S19\\

\newpage


\section*{SUPPLEMENTARY METHODS}

\subsection*{Mapping algorithm from vision-based tactile sensor to cutaneous shape display}

Spatial contact patterns were measured using a vision-based tactile sensor (GelSight Mini) and converted into binary actuation commands for a cutaneous shape display. To accomplish this, raw tactile sensor images were processed and mapped to localized inflation patterns using Algorithm~\ref{alg:mapping}. In our experiments, we used $r = 10$, $T_{\mathrm{noise\_floor}} = 800$, and $T_{\mathrm{contact}} = 1{,}200$. An overview of the mapping is shown in Fig.~\ref{fig:mapping}.

\begin{algorithm}
\caption{Tactile Sensor to Cutaneous Actuator Mapping}\label{alg:mapping}
\label{alg:mapping}
\begin{algorithmic}
\State Capture reference frame $I_\mathrm{0}$
\While{True}
    \State Capture frame $I$
    \State $I_\mathrm{d} \gets I - I_\mathrm{0}$
    \State $I_\mathrm{d} \gets \mathrm{Grayscale}(I_\mathrm{d})$
    \State $I_\mathrm{d} \gets \mathrm{GaussianBlur}(I_\mathrm{d})$
    \If{$\|I_\mathrm{d}\|_F > T_{\mathrm{noise\_floor}}$}
        \State $I_\mathrm{d} \gets \mathrm{OtsuThreshold}(I_\mathrm{d})$
    \Else
        \State $I_\mathrm{d} \gets \mathbf{0}_{m \times n}$
    \EndIf
    \For{each tactile pixel $i$ with parameters $(x_i, y_i, r)$}
        \State $E_i \gets I_\mathrm{d}[x_i - r : x_i + r,\ y_i - r : y_i + r]$
        \If{$\|E_i\|_F > T_{\mathrm{contact}}$}
            \State Inflate pixel $i$
        \Else
            \State Deflate pixel $i$
        \EndIf
    \EndFor
\EndWhile
\end{algorithmic}
\end{algorithm}

\begin{figure} [p]
	\centering
	\includegraphics[width=0.5\textwidth]{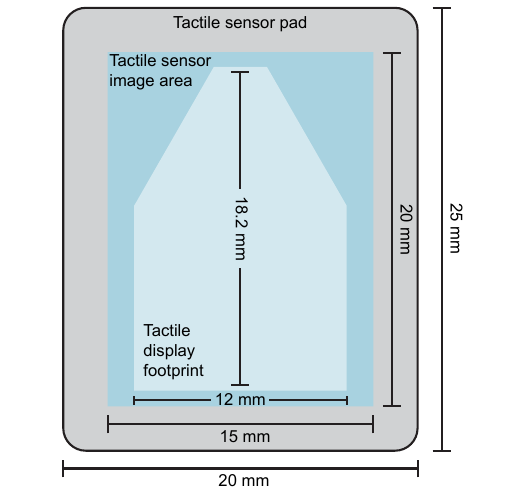}

    \singlespace
	\caption{\textbf{Comparison of tactile display and tactile sensor area.}
		The camera inside the vision-based tactile sensor only images a subset (blue region) of the tactile sensor's physical ``fingerpad" (gray region). The physical size of the tactile display (white region) is approximately equal to the imageable portion of the tactile sensing pad.
    }
	\label{fig:sensing_area}
\end{figure}

The physical ``fingerpad" of the tactile sensor used in the experiments measured $25\ \mathrm{mm} \times20\ \mathrm{mm}$ (Fig.~\ref{fig:sensing_area}, gray region). However, the sensor's camera captured only a $20\ \mathrm{mm} \times15\ \mathrm{mm}$ region of the fingerpad (Fig.~\ref{fig:sensing_area}, blue region), leaving the perimeter of the physical pad outside the sensor's detectable area. We found that this limitation did not substantially impair participants' ability to perform the tasks. The physical dimensions of the tactile display (Fig.~\ref{fig:sensing_area}, white region) closely matched the sensor's imageable area, leaving only 1.5\,mm of uncovered area on the left and right and less than 1\,mm along the top and bottom. Consequently, the distribution of contacts recorded by the tactile sensor was mapped approximately 1:1 onto the tactile display, without spatial stretching or compression, meaning that two points of contact 6\,mm apart on the tactile sensor would activate pixels 6\,mm apart on the tactile display. This 1:1 correspondence may have enabled participants to develop an intuitive understanding of the peg's relative position in the environment, thereby facilitating more optimal finger repositioning in the \textit{32D} condition. Had the mapping been spatially stretched or compressed, it might have distorted the operator's perception of finger position relative to the environment or required them to learn the altered spatial correspondence, though additional experiments would be required to validate this.

\subsection*{Experimental setup and data collection}

\begin{figure}[p]
	\centering
	\includegraphics[width=1.0\textwidth]{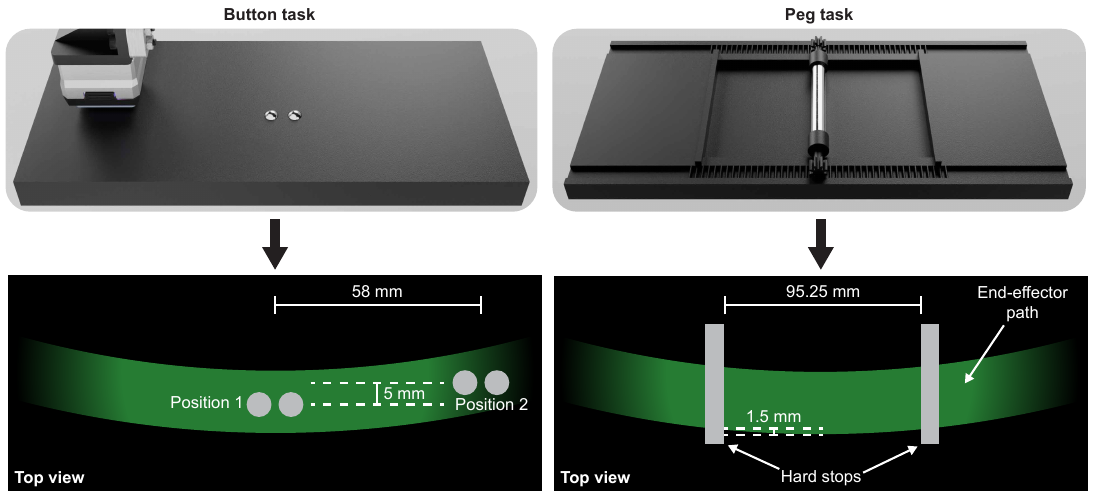}

    \singlespace
	\caption{\textbf{Task layout.}
		As the operator moved in the forward--backward direction, the end-effector followed a circular path (green) due to the robot's revolute joints. Because the button locations were shifted by 58\,mm between Position 1 and Position 2 in the button-discrimination task (left panel), they had to be offset by 5\,mm to accommodate this circular path. In the peg-rolling task (right), this circular path meant that the robot finger could slip laterally on the peg by a maximum of 1.5\,mm between the two hard stops. Relevant measurements are annotated; diagram is not drawn to scale.
    }
	\label{fig:task_layout}
\end{figure}

Participants completed tasks using a 2-DoF leader-follower telemanipulation system. Cutaneous feedback was provided via a spatially programmable fingertip display, comprising 32 independently inflatable elastic domes with a center-to-center pitch of 3\,mm. Kinesthetic force feedback was delivered through position-position bilateral control, capable of rendering end-effector stiffnesses of 9.4\,N/mm in the vertical direction and 2.2\,N/mm in the horizontal direction. The cutaneous feedback condition was varied across the trials (\textit{Off}, \textit{1D}, \textit{2D}, \textit{32D}), whereas kinesthetic feedback was active during all trials. In the button task, to make it more difficult for participants to memorize the positions of the two buttons, button locations were randomly alternated between two spatial configurations across trials. The inter-button spacing was 1\,cm in both configurations, while the button array was translated 58\,mm along the x-axis between layouts (Fig.~\ref{fig:task_layout}, left panel). All trajectories in Fig.~\ref{fig:buttons}B and Fig.~\ref{fig:dataset}A are from trials using the farther set of buttons (Fig.~\ref{fig:task_layout}, bottom left panel, Position 2).
 
We note that, because each DoF consists of a revolute joint, the end-effector technically moves along the surface of an ellipsoid. However, because the workspace was limited, we approximate this motion as planar for simplicity throughout the discussion. In the button-discrimination task, because the two button sets were shifted by 58\,mm between the two positions, they required a 5-mm offset to approximately align the buttons with the center of the robot finger at both locations (Fig.~\ref{fig:task_layout}, left panel). In the peg-rolling task, the arc traced by the tactile sensor could theoretically produce up to 1.5\,mm of lateral slip along the peg between the two hard stops (Fig.~\ref{fig:task_layout}, right panel). However, because the peg could not be rolled from one hard stop to another in a single stroke, the lateral slip during any individual rolling stroke was limited to approximately 1\,mm. We found that these minor slips did not appreciably affect the participants' ability to perform the task.

Prior to the teleoperated trials, participants performed the same task directly using the index finger of their dominant hand, and the recorded trajectories served as baselines for later comparison. For the baseline measurements, the tactile sensor was removed and the participant's finger was mounted to the follower device using a 3D-printed fixture and hook-and-loop strap that positioned the finger in roughly the same location as the tactile sensor. This provided a method of recording the participant's fingertip movements during direct manipulation while also allowing us to characterize natural task performance given the kinematic workspace constraints of the robot. Participants were also blindfolded during the baseline trials to match the absence of visual feedback in teleoperated conditions. For the button-discrimination task, each participant completed 12 direct-manipulation trials (3 repetitions per button and per button layout). For the peg-rolling task, each participant completed 6 direct-manipulation repetitions of the peg-rolling task using their bare finger.

During the teleoperated trials, participants' view of the task and robot end-effector was completely obstructed by a monitor displaying a graphical user interface (GUI). For both tasks, the GUI displayed the current trial number. In the button-discrimination task, the GUI instructed the participant on which of the two buttons to push during the current trial. The buttons were connected to a microcontroller (Teensy 4.1, SparkFun Electronics, Boulder, CO, USA), which communicated with the GUI to notify the participant when the correct button was pushed and the trial had ended. The trial would not end and the GUI would not notify the user if the incorrect button was pushed or if both buttons were pushed simultaneously. In the peg-rolling task, the GUI instructed the participant to ``Roll Forward" at the start of each trial. When the participant finished the forward roll by reaching the far hard stop, the experimenter would push a button that would change the GUI instructions to ``Roll Backward," notifying the participant to begin rolling the peg backward.

\subsection*{Trajectory preprocessing and alignment}

For the button task, direct-manipulation and teleoperated trajectories were spatially aligned prior to analysis. Direct trajectories were shifted such that their endpoints coincided with the mean button-press location across corresponding teleoperated trials (i.e., matched by target button and button location). All trajectories were then truncated to a common starting position (x = 92\,mm) to reduce variability due to differences in initial finger placement. For visualization in Fig.~\ref{fig:buttons}B, x-coordinates were offset by 92\,mm so that all trajectories originated at 0\,mm.

\subsection*{Behavioral metrics and similarity analysis}

We quantified the similarity between teleoperated and direct-manipulation trajectories using 2D dynamic time warping (DTW), which optimally aligns the trajectories in time and computes the distance between them~\cite{sakoe}. For each teleoperated trial, DTW distances were computed to all corresponding direct-manipulation trials (matched by target button and button location where applicable), and the resulting values were averaged to obtain a single DTW distance per trial.

\subsubsection*{Peg-rolling task metrics.}
For the peg-rolling task, in addition to computing the DTW distance between teleoperated and natural trajectories, we segmented trajectories into strokes, slips, loopbacks, and miscellaneous motions (Fig.~\ref{fig:peg_rolling}B). A stroke was defined as a continuous period of contact between the fingertip and the peg. Segments in which the participant continued to move past the peg were omitted from stroke length calculations. Additionally, the final stroke in either direction was occasionally short because the peg was near the end. Therefore, final strokes shorter than 10\,mm were excluded to reduce boundary effects. When counting the number of slips, we defined a slip as rolling past the peg and making contact with the surface beneath it. Contact with the surface was operationally defined as the fingertip descending below a y-coordinate of 9\,mm after rolling past the peg. Overshoot without surface contact was not counted as a slip.

\subsubsection*{Inter-trial variability analysis.}

In addition to using DTW distance to measure the deviation of teleoperated movements from natural human behaviors, we measured consistency within teleoperation datasets by computing the DTW distances between all pairs of teleoperated trials. We first computed the pairwise DTW distances between trajectories within each participant and feedback condition. For the button task, comparisons were restricted to trials with the same target button and button location (4 combinations). This yielded 12 pairwise comparisons per participant per condition, averaged for Fig.~\ref{fig:dataset}D. Then, the trials from all participants were combined to produce an aggregated dataset for each feedback condition, and DTW distances were computed between all pairs of trials in the aggregated datasets. Pairwise comparisons across participants yielded 630 distances per combination and feedback condition, or 2,520 distances per feedback condition (Fig.~\ref{fig:dataset}F). For the peg-rolling task, 66 pairwise comparisons were computed per participant per feedback condition and averaged for Fig.~\ref{fig:dataset}E. Pairwise comparisons within the aggregated datasets (i.e., across participants) yielded 7,140 distances per feedback condition (Fig.~\ref{fig:dataset}G).

\subsection*{Autonomous policy learning}

The $\pi_{0.5}$ policies were fine-tuned using the pre-trained \textit{pi05\_base} weights (Physical Intelligence, San Francisco, CA, USA) hosted on Hugging Face, with the LeRobot v0.6.1 software implementation. Because video data were recorded at 30 frames per second (fps), whereas leader and follower joint angles were recorded at 50\,Hz, joint-angle measurements were resampled to 30\,Hz prior to training. The external camera images ($1920 \times 1080$ pixels) were resized to $400 \times 224$, while images from the vision-based tactile sensor ($240 \times 320$ pixels) were used at their native resolution. Due to issues with collecting videos for the first two participants, the policies were trained only on data from participants 3 through 10 (96 trials per condition). Only the forward-rolling segment of each trial was used for training. For each condition, five trials were held out for validation, leaving 91 trials for training. The policies were trained for 3000 steps each on an NVIDIA RTX6000 GPU (batch size 32, AdamW optimizer, $2.5 \times 10^{-5}$ peak learning rate with cosine decay). The checkpoint with the best performance on the held-out trials was used for the autonomous rollouts (step 1500 for both).

During autonomous execution, the policies were run with five denoising steps on an NVIDIA GeForce RTX3090 GPU. Each inference required approximately 0.19\,s to generate a trajectory consisting of 50 leader-position setpoints. The predicted leader positions were provided as setpoints for the follower robot, which tracked them using a unilateral controller with the same gains as the position-position bilateral controller used in the behavioral experiments. To prevent discontinuities in the commanded trajectory that could arise from frequent replanning and setpoint updates, new trajectories were only generated 0.05\,s after the previous trajectory had finished executing. During the collection of autonomous demonstrations, a trial was classified as a ``failure" if the robot either began pushing the peg with the side of the tactile sensor rather than the bottom of the sensing pad, or applied an excessive force with the edge of its finger that caused the peg to shoot out from underneath it.


\section*{SUPPLEMENTARY TEXT}

\subsection*{Statistical results}

Tables S1--S19 contain the statistics and p-values for all metrics reported in the manuscript. All reported p-values were adjusted for multiple comparisons using the Holm-Bonferroni correction.

\subsubsection*{Button-discrimination task.}

In the button-discrimination task, \textit{32D} spatially distributed cutaneous feedback reduced the mean overshoot and mean number of switchbacks compared to the \textit{2D} (overshoot, $W = 0.0,\ p_{\mathrm{adj}} < 0.001$; switchbacks, $W = 0.0,\ p_{\mathrm{adj}} < 0.001$), \textit{1D} (overshoot, $W = 0.0,\ p_{\mathrm{adj}} < 0.001$; switchbacks, $W = 4.5,\ p_{\mathrm{adj}} = 0.004$), and \textit{Off} (overshoot, $W = 0.0,\ p_{\mathrm{adj}} < 0.001$; switchbacks, $W = 9.0,\ p_{\mathrm{adj}} = 0.007$) conditions (Tables~\ref{tab:buttons_overshoot} and \ref{tab:buttons_switchbacks}).

For DTW distance, the ANOVA revealed a highly significant main effect of Feedback Condition ($F(3,572) = 25.119,\ p < 0.001$; Fig.~\ref{fig:buttons}D). The post hoc pairwise comparisons found significant differences between the \textit{Off} condition and the \textit{1D} ($F(1,572) = 26.821,\ p_{\mathrm{adj}} < 0.001$), 2D ($F(1,572) = 47.100,\ p_{\mathrm{adj}} < 0.001$), and \textit{32D} ($F(1,572) = 64.471,\ p_{\mathrm{adj}} < 0.001$) conditions, suggesting that all levels of cutaneous feedback meaningfully improved task naturalness (Table~\ref{tab:buttons_dtw}). Furthermore, while no significant difference was found between \textit{1D} and \textit{2D} cutaneous feedback ($F(1,572) = 2.836,\ p_{\mathrm{adj}} = 0.185$), the highly localized, \textit{32D} feedback produced significantly lower DTW distances than the \textit{1D} feedback ($F(1,572) = 8.125,\ p_{\mathrm{adj}} = 0.014$). No other comparisons were significant. Wilcoxon signed-rank tests on the per-participant standard deviation of DTW distance (Fig.~\ref{fig:buttons}E) showed significantly lower variability in the \textit{32D} condition than in the \textit{2D} ($W = 11.0,\ p_{\mathrm{adj}} = 0.013$), \textit{1D} ($W = 0.0,\ p_{\mathrm{adj}} < 0.001$), and \textit{Off} ($W = 0.0,\ p_{\mathrm{adj}} < 0.001$) conditions (Table~\ref{tab:buttons_dtw_std}).

For task completion times, the ANOVA revealed a highly significant main effect of Feedback Condition ($F(3,572) = 26.598,\ p < 0.001$; Fig.~\ref{fig:buttons}F). The post hoc pairwise coefficient tests indicated that completion times were significantly reduced compared to the \textit{Off} condition for the \textit{1D} ($F(1,572) = 36.784,\ p_{\mathrm{adj}} < 0.001$), \textit{2D} ($F(1,572) = 45.621,\ p_{\mathrm{adj}} < 0.001$), and \textit{32D} ($F(1,572) = 69.188,\ p_{\mathrm{adj}} < 0.001$) conditions (Table~\ref{tab:buttons_time}). The difference between the \textit{1D} and \textit{32D} conditions did not reach significance after correction ($F(1,572) = 5.076,\ p = 0.025,\ p_{\mathrm{adj}} = 0.074$). Similarly to DTW distance, the differences in completion times between the \textit{1D} and \textit{2D} conditions ($F(1,572) = 0.475,\ p_{\mathrm{adj}} = 0.491$) and the \textit{2D} and \textit{32D} conditions ($F(1,572) = 2.445,\ p_{\mathrm{adj}} = 0.237$) were not significant. Wilcoxon signed-rank tests comparing the per-participant standard deviation of completion time revealed a significant difference between \textit{32D} feedback and the \textit{2D} ($W = 4.0,\ p_{\mathrm{adj}} = 0.003$), \textit{1D} ($W = 6.0,\ p_{\mathrm{adj}} = 0.003$), and \textit{Off} ($W = 0.0,\ p_{\mathrm{adj}} < 0.001$) conditions (Table~\ref{tab:buttons_time_std}).

Finally, the per-participant mean inter-trial DTW distance was significantly lower for the \textit{32D} condition when compared to trials where no cutaneous feedback was received ($W = 0.0,\ p_{\mathrm{adj}} < 0.001$; Table~\ref{tab:buttons_mean_pairwise}). Across all participants and trials, the Mann-Whitney \textit{U} tests revealed that the inter-trial DTW distances were significantly lower for the \textit{32D} condition than the \textit{2D} ($U = 2548888.0,\ p_{\mathrm{adj}} < 0.001$), \textit{1D} ($U = 2525668.0,\ p_{\mathrm{adj}} < 0.001$), and \textit{Off} ($U = 1235360.0,\ p_{\mathrm{adj}} < 0.001$) conditions (Table~\ref{tab:buttons_all_pairwise}).

\subsubsection*{Peg-rolling task.}

In the peg-rolling task, \textit{32D} spatially distributed cutaneous feedback increased the mean stroke length and reduced the number of strokes and slips off the peg compared to the \textit{2D} (stroke length, $W = 53.0,\ p_{\mathrm{adj}} = 0.004$; stroke count, $W = 6.0,\ p_{\mathrm{adj}} = 0.014$; slips, $W = 8.0,\ p_{\mathrm{adj}} = 0.039$), \textit{1D} (stroke length, $W = 54.0,\ p_{\mathrm{adj}} = 0.004$; stroke count, $W = 3.0,\ p_{\mathrm{adj}} = 0.010$; slips, $W = 1.5,\ p_{\mathrm{adj}} = 0.018$), and \textit{Off} (stroke length, $W = 55.0,\ p_{\mathrm{adj}} = 0.003$; stroke count, $W = 1.0,\ p_{\mathrm{adj}} = 0.006$; slips, $W = 5.0,\ p_{\mathrm{adj}} = 0.039$) conditions (Tables~\ref{tab:peg_slips}--\ref{tab:peg_strokes}).

For DTW distance, the ANOVA revealed a highly significant main effect of Feedback Condition ($F(3,476) = 18.16,\ p < 0.001$; Fig.~\ref{fig:peg_rolling}E). Similarly to the button-discrimination task, post hoc pairwise comparisons found a significant difference between the \textit{Off} condition and the \textit{2D} ($F(1,476) = 14.861,\ p_{\mathrm{adj}} < 0.001$) and \textit{32D} ($F(1,476) = 51.071,\ p_{\mathrm{adj}} < 0.001$) conditions (Table~\ref{tab:peg_dtw}). Although no significant difference was found between \textit{1D} and \textit{2D} cutaneous feedback ($F(1,476) = 2.794,\ p_{\mathrm{adj}} = 0.095$) or \textit{1D} and no (\textit{Off}) cutaneous feedback ($F(1,476) = 4.767,\ p_{\mathrm{adj}} = 0.059$), the highly localized, \textit{32D} feedback produced significantly lower DTW distances than \textit{2D} feedback ($F(1,476) = 10.833,\ p_{\mathrm{adj}} = 0.003$) and \textit{1D} feedback ($F(1,476) = 24.632,\ p_{\mathrm{adj}} < 0.001$). Wilcoxon signed-rank tests on the per-participant standard deviation of DTW distance revealed significantly lower variability in the \textit{32D} condition than in the \textit{2D} ($W = 0.0,\ p_{\mathrm{adj}} = 0.003$), \textit{1D} ($W = 3.0,\ p_{\mathrm{adj}} = 0.005$), and \textit{Off} ($W = 0.0,\ p_{\mathrm{adj}} = 0.003$) conditions (Fig.~\ref{fig:peg_rolling}F; Table~\ref{tab:peg_dtw_std}).

For task completion times, the ANOVA revealed a highly significant main effect of Feedback Condition ($F(3,476) = 40.727,\ p < 0.001$; Fig.~\ref{fig:peg_rolling}G). Post hoc pairwise coefficient tests indicated that completion times were significantly reduced compared to the \textit{Off} condition for the \textit{2D} ($F(1,476) = 24.969,\ p_{\mathrm{adj}} < 0.001$) and \textit{32D} ($F(1,476) = 104.059,\ p_{\mathrm{adj}} < 0.001$) conditions, but not for the \textit{1D} condition ($F(1,476) = 2.533,\ p_{\mathrm{adj}} = 0.112$; Table~\ref{tab:peg_time}). The completion times for the \textit{32D} condition were also significantly shorter than those of the \textit{2D} ($F(1,476) = 27.082,\ p_{\mathrm{adj}} < 0.001$) and \textit{1D} conditions ($F(1,476) = 74.124,\ p_{\mathrm{adj}} < 0.001$). The \textit{2D} condition also led to significantly shorter completion times than the \textit{1D} condition ($F(1,476) = 11.597,\ p_{\mathrm{adj}} = 0.001$). Wilcoxon signed-rank tests on the per-participant standard deviation of completion time revealed significantly lower variability in the \textit{32D} condition than in the \textit{2D} ($W = 0.0,\ p_{\mathrm{adj}} = 0.003$), \textit{1D} ($W = 0.0,\ p_{\mathrm{adj}} = 0.003$), and \textit{Off} ($W = 0.0,\ p_{\mathrm{adj}} = 0.003$) conditions (Fig.~\ref{fig:peg_rolling}H; Table~\ref{tab:peg_time_std}).

Finally, the per-participant mean inter-trial DTW distance was significantly lower for the \textit{32D} condition when compared to each of the other feedback conditions ($W = 0.0,\ p_{\mathrm{adj}} = 0.003$; Table~\ref{tab:peg_mean_pairwise}). Across all participants and trials, the Mann-Whitney \textit{U} tests revealed that the inter-trial DTW distances were significantly lower for the \textit{32D} condition than the \textit{2D} ($U = 9220395.0,\ p_{\mathrm{adj}} < 0.001$), \textit{1D} ($U = 4593005.0,\ p_{\mathrm{adj}} < 0.001$), and \textit{Off} ($U = 4261698.0,\ p_{\mathrm{adj}} < 0.001$) conditions (Table~\ref{tab:peg_all_pairwise}).

\subsubsection*{Autonomous robot policies.}

Autonomous trajectories generated by the \textit{32D} policy exhibited significantly lower DTW distances to direct human trajectories than those generated by the \textit{Off} policy (31\% lower on average; $U = 6.0,\ p < 0.001$; Fig.~\ref{fig:autonomy}B). The \textit{32D} policy also achieved significantly shorter task completion times than the \textit{Off} policy (29\% faster on average; $U = 16.0,\ p < 0.001$; Fig.~\ref{fig:autonomy}C).

\clearpage


\section*{SUPPLEMENTARY FIGURES AND TABLES}

\makeatletter
\setlength{\@fptop}{0pt plus 1fil}   
\setlength{\@fpsep}{12pt plus 1fil}  
\setlength{\@fpbot}{0pt plus 1fil}   
\makeatother

\vspace*{20mm}
\begin{figure}[h]
	\centering
	\includegraphics[width=0.5\textwidth]{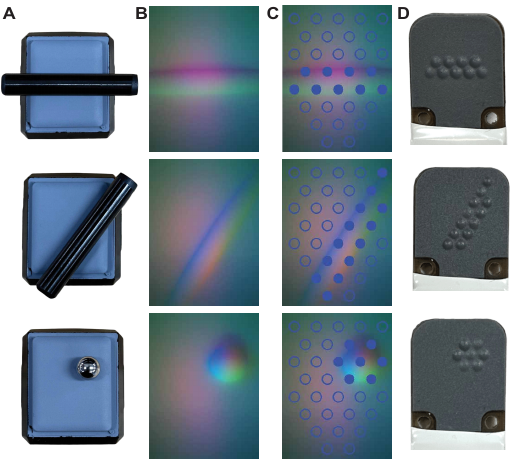}

    \singlespace
	\caption{\textbf{Mapping from vision-based tactile sensor to cutaneous shape display.}
		(\textbf{A}) The vision-based tactile sensor makes contact with objects in the environment.
        (\textbf{B}) The vision-based tactile sensor outputs a $240 \times 320$ RGB image.
        (\textbf{C}) A mapping algorithm converts the raw RGB sensor image into a pixel inflation pattern (solid blue circles) that mirrors the pixel layout of the cutaneous shape display (blue circles).
        (\textbf{D}) The corresponding pixels on the cutaneous shape display are inflated.
    }
	\label{fig:mapping}
\end{figure}

\vspace*{35mm}
\begin{figure}[h]
	\centering
	\includegraphics[width=1.0\textwidth]{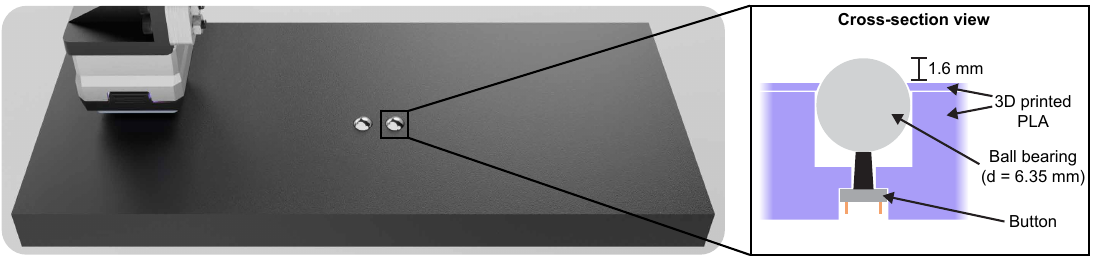}

    \singlespace
	\caption{\textbf{Construction of ball-bearing buttons.}
		Each button consists of a 6.35-mm-diameter ball bearing constrained by a 3D-printed PLA piece on top and a breadboard button on the bottom. The ball-bearing button protrudes 1.6\,mm from the top surface and requires a peak applied force of 2.7\,N and a displacement of 0.25\,mm to register a click.
    }
	\label{fig:ball_bearing_button}
\end{figure}

\clearpage

\begin{table}
	\centering
	
	\caption{\textbf{Button-Discrimination Task: Overshoot Distance.}
		Pairwise comparisons of participants' mean overshoot distance between the \textit{32D} condition and the \textit{2D}, \textit{1D}, and \textit{Off} conditions. Statistics are from one-tailed Wilcoxon signed-rank tests.}
	\label{tab:buttons_overshoot}

    \vspace{12pt}
    
    \begin{tabular}{lccc}
    \noalign{\hrule height 1pt}
                            & \textbf{\textit{2D}}                                   & \textbf{\textit{1D}}                                   & \textbf{\textit{Off}} \\
    \noalign{\hrule height 1pt}
    \textbf{\textit{32D}}           & \makecell{$W = 0.0$\\$p < 0.001^{***}$}       & \makecell{$W = 0.0$\\$p < 0.001^{***}$}       & \makecell{$W = 0.0$\\$p < 0.001^{***}$}
    \end{tabular}
\end{table}

\begin{table}
	\centering
	
	\caption{\textbf{Button-Discrimination Task: Number of Switchbacks.}
		Pairwise comparisons of participants' mean number of switchbacks between the \textit{32D} condition and the \textit{2D}, \textit{1D}, and \textit{Off} conditions. Statistics are from one-tailed Wilcoxon signed-rank tests.}
	\label{tab:buttons_switchbacks}

    \vspace{12pt}
    
    \begin{tabular}{lccc}
    \noalign{\hrule height 1pt}
                            & \textbf{\textit{2D}}                                   & \textbf{\textit{1D}}                                   & \textbf{\textit{Off}} \\
    \noalign{\hrule height 1pt}
    \textbf{\textit{32D}}           & \makecell{$W = 0.0$\\$p < 0.001^{***}$}       & \makecell{$W = 4.5$\\$p = 0.004^{**}$}        & \makecell{$W = 9.0$\\$p = 0.007^{**}$}
    \end{tabular}
\end{table}

\begin{table}
	\centering
	
	\caption{\textbf{Button-Discrimination Task: Trajectory Naturalness (DTW Distance).}
		Pairwise comparisons of DTW distance (normalized by each participant's mean DTW distance across all trials) between all six feedback-condition pairs. Statistics are from two-tailed coefficient tests.}
	\label{tab:buttons_dtw}

    \vspace{12pt}
    
    \begin{tabular}{lccc}
    \noalign{\hrule height 1pt}
                            & \textbf{\textit{2D}}                                   & \textbf{\textit{1D}}                                       & \textbf{\textit{Off}} \\
    \noalign{\hrule height 1pt}
    \textbf{\textit{32D}}           & \makecell{$F(1,572) = 1.361$\\$p = 0.244$}    & \makecell{$F(1,572) = 8.125$\\$p = 0.014^{*}$}    & \makecell{$F(1,572) = 64.471$\\$p < 0.001^{***}$} \\
    \hline
    \textbf{\textit{2D}}             &                                               & \makecell{$F(1,572) = 2.836$\\$p = 0.185$}        & \makecell{$F(1,572) = 47.100$\\$p < 0.001^{***}$} \\
    \hline
    \textbf{\textit{1D}}             &                                               &                                  & \makecell{$F(1,572) = 26.821$\\$p < 0.001^{***}$}
    \end{tabular}
\end{table}

\begin{table}
	\centering
	
	\caption{\textbf{Button-Discrimination Task: Variability in Trajectory Naturalness.}
		Pairwise comparisons of within-participant standard deviation of DTW distance between the \textit{32D} condition and the \textit{2D}, \textit{1D}, and \textit{Off} conditions. Statistics are from one-tailed Wilcoxon signed-rank tests.}
	\label{tab:buttons_dtw_std}

    \vspace{12pt}
    
    \begin{tabular}{lccc}
    \noalign{\hrule height 1pt}
                            & \textbf{\textit{2D}}                                   & \textbf{\textit{1D}}                                   & \textbf{\textit{Off}} \\
    \noalign{\hrule height 1pt}
    \textbf{\textit{32D}}           & \makecell{$W = 11.0$\\$p = 0.013^{*}$}        & \makecell{$W = 0.0$\\$p < 0.001^{***}$}       & \makecell{$W = 0.0$\\$p < 0.001^{***}$}
    \end{tabular}
\end{table}

\begin{table}
	\centering
	
	\caption{\textbf{Button-Discrimination Task: Completion Time.}
		Pairwise comparisons of task completion time (normalized by each participant's mean completion time across all trials) between all six feedback-condition pairs. Statistics are from two-tailed coefficient tests.}
	\label{tab:buttons_time}

    \vspace{12pt}
    
    \begin{tabular}{lccc}
    \noalign{\hrule height 1pt}
                            & \textbf{\textit{2D}}                                   & \textbf{\textit{1D}}                                       & \textbf{\textit{Off}} \\
    \noalign{\hrule height 1pt}
    \textbf{\textit{32D}}           & \makecell{$F(1,572) = 2.445$\\$p = 0.237$}    & \makecell{$F(1,572) = 5.076$\\$p = 0.074$}        & \makecell{$F(1,572) = 69.188$\\$p < 0.001^{***}$} \\
    \hline
    \textbf{\textit{2D}}             &                                               & \makecell{$F(1,572) = 0.475$\\$p = 0.491$}        & \makecell{$F(1,572) = 45.621$\\$p < 0.001^{***}$} \\
    \hline
    \textbf{\textit{1D}}             &                                               &                                                   & \makecell{$F(1,572) = 36.784$\\$p < 0.001^{***}$}
    \end{tabular}
\end{table}

\begin{table}
	\centering
	
	\caption{\textbf{Button-Discrimination Task: Variability in Completion Time.}
		Pairwise comparisons of within-participant standard deviation of completion time between the \textit{32D} condition and the \textit{2D}, \textit{1D}, and \textit{Off} conditions. Statistics are from one-tailed Wilcoxon signed-rank tests.}
	\label{tab:buttons_time_std}

    \vspace{12pt}
    
    \begin{tabular}{lccc}
    \noalign{\hrule height 1pt}
                            & \textbf{\textit{2D}}                                   & \textbf{\textit{1D}}                                   & \textbf{\textit{Off}} \\
    \noalign{\hrule height 1pt}
    \textbf{\textit{32D}}           & \makecell{$W = 4.0$\\$p = 0.003^{**}$}        & \makecell{$W = 6.0$\\$p = 0.003^{**}$}        & \makecell{$W = 0.0$\\$p < 0.001^{***}$}
    \end{tabular}
\end{table}

\begin{table}
	\centering
	
	\caption{\textbf{Peg-Rolling Task: Slip Frequency.}
		Pairwise comparisons of participants' mean number of slips per trial between the \textit{32D} condition and the \textit{2D}, \textit{1D}, and \textit{Off} conditions. Statistics are from one-tailed Wilcoxon signed-rank tests.}
	\label{tab:peg_slips}

    \vspace{12pt}
    
    \begin{tabular}{lccc}
    \noalign{\hrule height 1pt}
                            & \textbf{\textit{2D}}                              & \textbf{\textit{1D}}                                & \textbf{\textit{Off}} \\
    \noalign{\hrule height 1pt}
    \textbf{\textit{32D}}           & \makecell{$W = 8.0$\\$p = 0.039^{*}$}    & \makecell{$W = 1.5$\\$p = 0.018^{*}$}      & \makecell{$W = 5.0$\\$p = 0.039^{*}$}
    \end{tabular}
\end{table}

\begin{table}
	\centering
	
	\caption{\textbf{Peg-Rolling Task: Stroke Length.}
		Pairwise comparisons of participants' mean stroke length between the \textit{32D} condition and the \textit{2D}, \textit{1D}, and \textit{Off} conditions. Statistics are from one-tailed Wilcoxon signed-rank tests.}
	\label{tab:peg_stroke_length}

    \vspace{12pt}
    
    \begin{tabular}{lccc}
    \noalign{\hrule height 1pt}
                            & \textbf{\textit{2D}}                                & \textbf{\textit{1D}}                                  & \textbf{\textit{Off}} \\
    \noalign{\hrule height 1pt}
    \textbf{\textit{32D}}           & \makecell{$W = 53.0$\\$p = 0.004^{**}$}    & \makecell{$W = 54.0$\\$p = 0.004^{**}$}      & \makecell{$W = 55.0$\\$p = 0.003^{**}$}
    \end{tabular}
\end{table}

\begin{table}
	\centering
	
	\caption{\textbf{Peg-Rolling Task: Number of Strokes.}
		Pairwise comparisons of participants' mean number of strokes between the \textit{32D} condition and the \textit{2D}, \textit{1D}, and \textit{Off} conditions. Statistics are from one-tailed Wilcoxon signed-rank tests.}
	\label{tab:peg_strokes}

    \vspace{12pt}
    
    \begin{tabular}{lccc}
    \noalign{\hrule height 1pt}
                            & \textbf{\textit{2D}}                                   & \textbf{\textit{1D}}                                   & \textbf{\textit{Off}} \\
    \noalign{\hrule height 1pt}
    \textbf{\textit{32D}}           & \makecell{$W = 6.0$\\$p = 0.014^{*}$}    & \makecell{$W = 3.0$\\$p = 0.010^{*}$}      & \makecell{$W = 1.0$\\$p = 0.006^{**}$}
    \end{tabular}
\end{table}

\begin{table}
	\centering
	
	\caption{\textbf{Peg-Rolling Task: Trajectory Naturalness (DTW Distance).}
		Pairwise comparisons of DTW distance (normalized by each participant's mean DTW distance across all trials) between all six feedback-condition pairs. Statistics are from two-tailed coefficient tests.}
	\label{tab:peg_dtw}

    \vspace{12pt}
    
    \begin{tabular}{lccc}
    \noalign{\hrule height 1pt}
                            & \textbf{\textit{2D}}                                           & \textbf{\textit{1D}}                                           & \textbf{\textit{Off}} \\
    \noalign{\hrule height 1pt}
    \textbf{\textit{32D}}           & \makecell{$F(1,476) = 10.833$\\$p = 0.003^{**}$}      & \makecell{$F(1,476) = 24.632$\\$p < 0.001^{***}$}     & \makecell{$F(1,476) = 51.071$\\$p < 0.001^{***}$} \\
    \hline
    \textbf{\textit{2D}}             &                                                       & \makecell{$F(1,476) = 2.794$\\$p = 0.095$}            & \makecell{$F(1,476) = 14.861$\\$p < 0.001^{***}$} \\
    \hline
    \textbf{\textit{1D}}             &                                                       &                                                       & \makecell{$F(1,476) = 4.767$\\$p = 0.059$}
    \end{tabular}
\end{table}

\begin{table}
	\centering
	
	\caption{\textbf{Peg-Rolling Task: Variability in Trajectory Naturalness.}
		Pairwise comparisons of within-participant standard deviation of DTW distance between the \textit{32D} condition and the \textit{2D}, \textit{1D}, and \textit{Off} conditions. Statistics are from one-tailed Wilcoxon signed-rank tests.}
	\label{tab:peg_dtw_std}

    \vspace{12pt}
    
    \begin{tabular}{lccc}
    \noalign{\hrule height 1pt}
                            & \textbf{\textit{2D}}                                   & \textbf{\textit{1D}}                                   & \textbf{\textit{Off}} \\
    \noalign{\hrule height 1pt}
    \textbf{\textit{32D}}           & \makecell{$W = 0.0$\\$p = 0.003^{**}$}        & \makecell{$W = 3.0$\\$p = 0.005^{**}$}        & \makecell{$W = 0.0$\\$p = 0.003^{**}$}
    \end{tabular}
\end{table}

\begin{table}
	\centering
	
	\caption{\textbf{Peg-Rolling Task: Completion Time.}
		Pairwise comparisons of task completion time (normalized by each participant's mean completion time across all trials) between all six feedback-condition pairs. Statistics are from two-tailed coefficient tests.}
	\label{tab:peg_time}

    \vspace{12pt}
    
    \begin{tabular}{lccc}
    \noalign{\hrule height 1pt}
                            & \textbf{\textit{2D}}                                           & \textbf{\textit{1D}}                                               & \textbf{\textit{Off}} \\
    \noalign{\hrule height 1pt}
    \textbf{\textit{32D}}           & \makecell{$F(1,476) = 27.082$\\$p < 0.001^{***}$}     & \makecell{$F(1,476) = 74.124$\\$p < 0.001^{***}$}         & \makecell{$F(1,476) = 104.059$\\$p < 0.001^{***}$} \\
    \hline
    \textbf{\textit{2D}}             &                                                       & \makecell{$F(1,476) = 11.597$\\$p = 0.001^{***}$}         & \makecell{$F(1,476) = 24.969$\\$p < 0.001^{***}$} \\
    \hline
    \textbf{\textit{1D}}             &                                                       &                                                           & \makecell{$F(1,476) = 2.533$\\$p = 0.112$}
    \end{tabular}
\end{table}

\begin{table}
	\centering
	
	\caption{\textbf{Peg-Rolling Task: Variability in Completion Time.}
		Pairwise comparisons of within-participant standard deviation of completion time between the \textit{32D} condition and the \textit{2D}, \textit{1D}, and \textit{Off} conditions. Statistics are from one-tailed Wilcoxon signed-rank tests.}
	\label{tab:peg_time_std}

    \vspace{12pt}
    
    \begin{tabular}{lccc}
    \noalign{\hrule height 1pt}
                            & \textbf{\textit{2D}}                                & \textbf{\textit{1D}}                                  & \textbf{\textit{Off}} \\
    \noalign{\hrule height 1pt}
    \textbf{\textit{32D}}           & \makecell{$W = 0.0$\\$p = 0.003^{**}$}    & \makecell{$W = 0.0$\\$p = 0.003^{**}$}      & \makecell{$W = 0.0$\\$p = 0.003^{**}$}
    \end{tabular}
\end{table}

\begin{table}
	\centering
	
	\caption{\textbf{Button-Discrimination Task: Questionnaire Responses.}
		Pairwise comparisons of post-experiment questionnaire responses between adjacent feedback conditions (\textit{Off} vs. \textit{1D}, \textit{1D} vs. \textit{2D}, and \textit{2D} vs. \textit{32D}). Statistics are from one-tailed Wilcoxon signed-rank tests.}
	\label{tab:buttons_questionnaire}

    \vspace{12pt}

    \begin{tabular}{lccc}
    \noalign{\hrule height 1pt}
                            & \textbf{\textit{Off} $\ \xrightarrow{}$ \textit{1D}}        & \textbf{\textit{1D} $\ \xrightarrow{}$ \textit{2D}}             & \textbf{\textit{2D} $\ \xrightarrow{}$ \textit{32D}} \\
    \noalign{\hrule height 1pt}
    \textbf{Q1}           & \makecell{$W = 59.0$\\$p = 0.022^{*}$}      & \makecell{$W = 51.0$\\$p = 0.022^{*}$}        & \makecell{$W = 25.0$\\$p = 0.047^{*}$} \\
    \hline
    \textbf{Q2}           & \makecell{$W = 78.0$\\$p < 0.001^{***}$}    & \makecell{$W = 41.0$\\$p = 0.031^{*}$}        & \makecell{$W = 28.0$\\$p = 0.117$} \\
    \hline
    \textbf{Q3}           & \makecell{$W = 62.5$\\$p = 0.010^{*}$}      & \makecell{$W = 28.0$\\$p = 0.016^{*}$}        & \makecell{$W = 21.5$\\$p = 0.141$} \\
    \hline
    \textbf{Q4}           & \makecell{$W = 0.0$\\$p < 0.001^{***}$}     & \makecell{$W = 0.0$\\$p = 0.016^{*}$}         & \makecell{$W = 7.5$\\$p = 0.094$}
    \end{tabular}
\end{table}

\begin{table}
	\centering
	
	\caption{\textbf{Peg-Rolling Task: Questionnaire Responses.}
		Pairwise comparisons of post-experiment questionnaire responses between adjacent feedback conditions (\textit{Off} vs. \textit{1D}, \textit{1D} vs. \textit{2D}, and \textit{2D} vs. \textit{32D}). Statistics are from one-tailed Wilcoxon signed-rank tests.}
	\label{tab:peg_questionnaire}

    \vspace{12pt}
    
    \begin{tabular}{lccc}
    \noalign{\hrule height 1pt}
                            & \textbf{\textit{Off} $\ \xrightarrow{}$ \textit{1D}}        & \textbf{\textit{1D} $\ \xrightarrow{}$ \textit{2D}}         & \textbf{\textit{2D} $\ \xrightarrow{}$ \textit{32D}} \\
    \noalign{\hrule height 1pt}
    \textbf{Q1}             & \makecell{$W = 21.0$\\$p = 0.047^{*}$}    & \makecell{$W = 40.5$\\$p = 0.047^{*}$}    & \makecell{$W = 30.5$\\$p = 0.047^{*}$} \\
    \hline
    \textbf{Q2}             & \makecell{$W = 15.0$\\$p = 0.063$}        & \makecell{$W = 36.0$\\$p = 0.012^{*}$}    & \makecell{$W = 25.0$\\$p = 0.063$} \\
    \hline
    \textbf{Q3}             & \makecell{$W = 10.0$\\$p = 0.125$}        & \makecell{$W = 15.0$\\$p = 0.094$}        & \makecell{$W = 18.5$\\$p = 0.125$} \\
    \hline
    \textbf{Q4}             & \makecell{$W = 6.0$\\$p = 0.266$}         & \makecell{$W = 2.5$\\$p = 0.059$}         & \makecell{$W = 6.0$\\$p = 0.250$} \\
    \end{tabular}
\end{table}

\begin{table}
	\centering
	
	\caption{\textbf{Button-Discrimination Task: Per-Participant Inter-Trial Trajectory Variability.}
		Pairwise comparisons  of participants' mean inter-trial DTW distance between the \textit{32D} condition and the \textit{2D}, \textit{1D}, and \textit{Off} conditions. Statistics are from one-tailed Wilcoxon signed-rank tests.}
	\label{tab:buttons_mean_pairwise}

    \vspace{12pt}
    
    \begin{tabular}{lccc}
    \noalign{\hrule height 1pt}
                            & \textbf{\textit{2D}}                                   & \textbf{\textit{1D}}                                   & \textbf{\textit{Off}} \\
    \noalign{\hrule height 1pt}
    \textbf{\textit{32D}}           & \makecell{$W = 19.0$\\$p = 0.065$}       & \makecell{$W = 15.0$\\$p = 0.065$}        & \makecell{$W = 0.0$\\$p < 0.001^{***}$}
    \end{tabular}
\end{table}

\begin{table}
	\centering
	
	\caption{\textbf{Peg-Rolling Task: Per-Participant Inter-Trial Trajectory Variability.}
		Pairwise comparisons  of participants' mean inter-trial DTW distance between the \textit{32D} condition and the \textit{2D}, \textit{1D}, and \textit{Off} conditions. Statistics are from one-tailed Wilcoxon signed-rank tests.}
	\label{tab:peg_mean_pairwise}

    \vspace{12pt}
    
    \begin{tabular}{lccc}
    \noalign{\hrule height 1pt}
                            & \textbf{\textit{2D}}                                   & \textbf{\textit{1D}}                                   & \textbf{\textit{Off}} \\
    \noalign{\hrule height 1pt}
    \textbf{\textit{32D}}           & \makecell{$W = 0.0$\\$p = 0.003^{**}$}       & \makecell{$W = 0.0$\\$p = 0.003^{**}$}        & \makecell{$W = 0.0$\\$p = 0.003^{**}$}
    \end{tabular}
\end{table}

\begin{table}
	\centering
	
	\caption{\textbf{Button-Discrimination Task: Aggregated Inter-Trial Trajectory Variability.}
		Pairwise comparisons of all pairwise inter-trial DTW distances between the \textit{32D} condition and the \textit{2D}, \textit{1D}, and \textit{Off} conditions. Statistics are from one-tailed Mann-Whitney \textit{U} tests.}
	\label{tab:buttons_all_pairwise}

    \vspace{12pt}
    
    \begin{tabular}{lccc}
    \noalign{\hrule height 1pt}
                            & \textbf{\textit{2D}}                                   & \textbf{\textit{1D}}                                   & \textbf{\textit{Off}} \\
    \noalign{\hrule height 1pt}
    \textbf{\textit{32D}}           & \makecell{$U = 2548888.0$\\$p < 0.001^{***}$}       & \makecell{$U = 2525668.0$\\$p < 0.001^{***}$}        & \makecell{$U = 1235360.0$\\$p < 0.001^{***}$}
    \end{tabular}
\end{table}

\begin{table}
	\centering
	
	\caption{\textbf{Peg-Rolling Task: Aggregated Inter-Trial Trajectory Variability.}
		Pairwise comparisons of all pairwise inter-trial DTW distances between the \textit{32D} condition and the \textit{2D}, \textit{1D}, and \textit{Off} conditions. Statistics are from one-tailed Mann-Whitney \textit{U} tests.}
	\label{tab:peg_all_pairwise}

    \vspace{12pt}
    
    \begin{tabular}{lccc}
    \noalign{\hrule height 1pt}
                            & \textbf{\textit{2D}}                                   & \textbf{\textit{1D}}                                   & \textbf{\textit{Off}} \\
    \noalign{\hrule height 1pt}
    \textbf{\textit{32D}}           & \makecell{$U = 9220395.0$\\$p < 0.001^{***}$}       & \makecell{$U = 4593005.0$\\$p < 0.001^{***}$}        & \makecell{$U = 4261698.0$\\$p < 0.001^{***}$}
    \end{tabular}
\end{table}

\end{document}